\documentclass[runningheads]{llncs}

\usepackage{eccv}

\usepackage{eccvabbrv}
\usepackage{adjustbox}

\usepackage{graphicx}
\usepackage{booktabs}
\usepackage{xcolor} 

\usepackage{wrapfig}

\usepackage[accsupp]{axessibility}  

\usepackage{hyperref}

\usepackage{orcidlink}

\begin{document}

\title{PreciseControl: Enhancing Text-To-Image Diffusion Models with Fine-Grained Attribute Control} 

\titlerunning{PreciseControl} 

\author{Rishubh Parihar\thanks{Equal contribution}\inst{1}\and
Sachidanand VS$^*$\inst{1} \and
Sabariswaran Mani\inst{2}\thanks{Work done during internship at VAL, IISc} \and 
Tejan Karmali\inst{1} \and
R. Venkatesh Babu\inst{1}}

\authorrunning{R. Parihar \textit{et al.}}

\institute{Vision and AI Lab, IISc Bangalore\and
IIT Kharagpur \\ 
\href{https://rishubhpar.github.io/PreciseControl.home}{Project Page}}

\maketitle
\begin{abstract}
Recently, we have seen a surge of personalization methods for text-to-image (T2I) diffusion models to learn a concept using a few images. Existing approaches, when used for face personalization, suffer to achieve convincing inversion with identity preservation and rely on semantic text-based editing of the generated face. However, a more fine-grained control is desired for facial attribute editing, which is challenging to achieve solely with text prompts. In contrast, StyleGAN models learn a rich face prior and enable smooth control towards fine-grained attribute editing by latent manipulation. This work uses the disentangled $\mathcal{W+}$ space of StyleGANs to condition the T2I model. This approach allows us to precisely manipulate facial attributes, such as smoothly introducing a smile, while preserving the existing coarse text-based control inherent in T2I models. To enable conditioning of the T2I model on the $\mathcal{W+}$ space, we train a latent mapper to translate latent codes from $\mathcal{W+}$ to the token embedding space of the T2I model. The proposed approach excels in the precise inversion of face images with attribute preservation and facilitates continuous control for fine-grained attribute editing. Furthermore, our approach can be readily extended to generate compositions involving multiple individuals. We perform extensive experiments to validate our method for face personalization and fine-grained attribute editing. 
\end{abstract}
\section{Introduction}
\begin{figure}[t]
    \centering
    \vspace{-3mm}
    \includegraphics[width=0.99\textwidth]{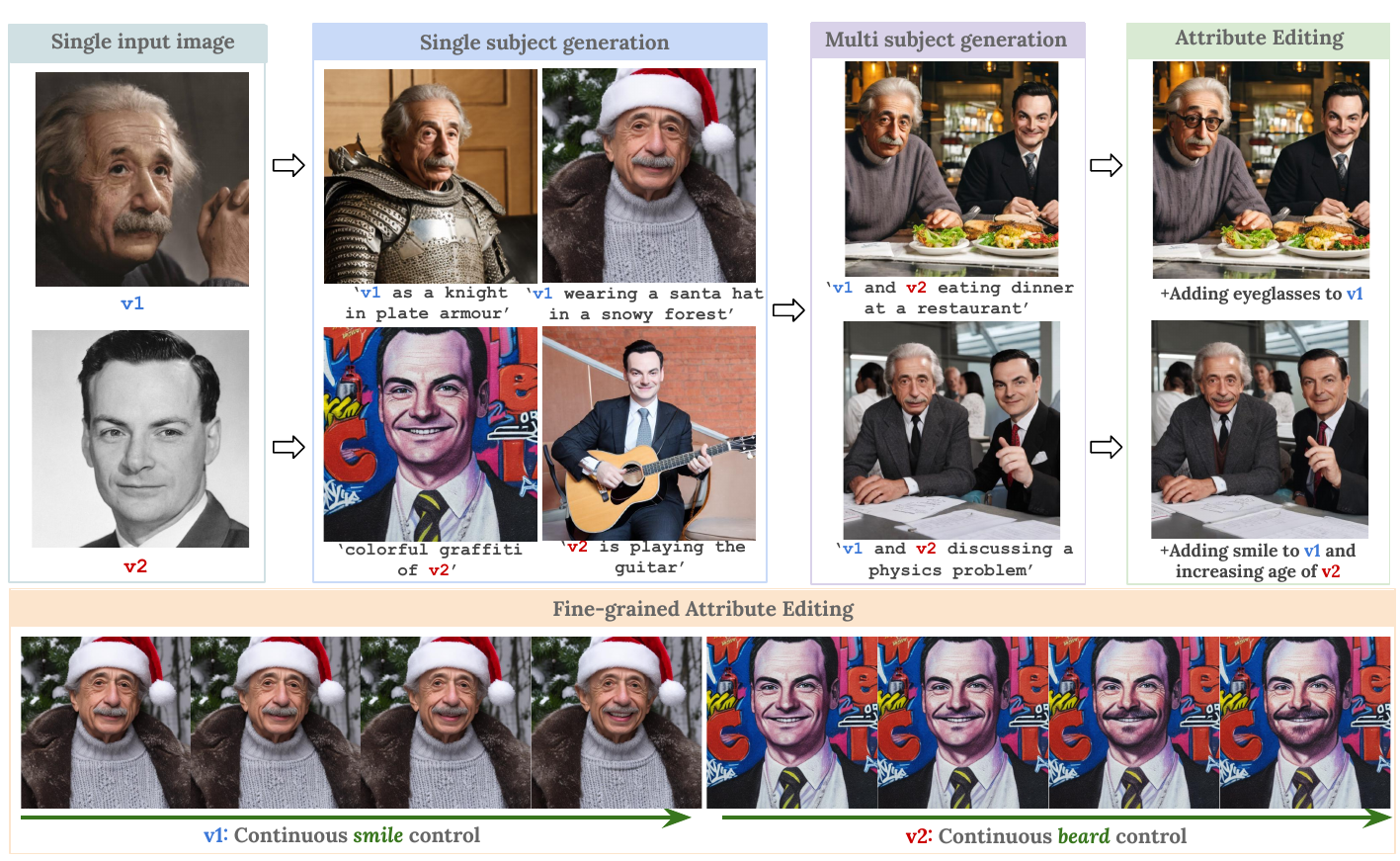}
    \vspace{-3mm}
    \caption{Given a single portrait image, we embed the subject into a text-to-image diffusion model for personalized image generation. The embedded subject can then be transformed or placed in a novel context using text conditioning. The proposed method can also compose multiple learned subjects with high fidelity and identity preservation. To obtain precise inversion of face, we condition the T2I model on the rich $\mathcal{W+}$ latent space of StyleGAN2. This enables our method to additionally perform fine-grained control over the generated face with continuous control over facial attributes such as age and beard.}
    \vspace{-5mm}
    \label{fig:teaser}
\end{figure}

Recent personalization methods~\cite{textual-inversion,ruiz2023dreambooth} for large text-to-image (T2I) diffusion models~\cite{ldm,imagen} aim to learn a new concept (e.g., your pet) given a few input images. The learned concept is 
then generated using text prompts in novel contexts (e.g. diverse backgrounds and poses) and styles, thus controlling coarse aspects of an image. Personalization of human portraits ~\cite{celeb-basis} is especially interesting due to the wide range of applications in entertainment and advertising. However, embedding faces into a generative model has its unique challenges, including faithful inversion of the subject's identity along with its fine facial features. More importantly, smooth control over facial attributes is crucial for precise editing of generated faces, which is challenging to achieve with only text (e.g., continuous increase in smile in Fig.\ref{fig:teaser}). 

Advancements in StyleGAN models~\cite{sg,sg2} have enabled the generation of highly realistic face images by learning a rich prior over face images. Further, these models have semantically meaningful and disentangled $\mathcal{W+}$ latent space ~\cite{interface} that enable fine-grained attribute control in the generated images ~\cite{harkonen2020ganspace,patashnik2021styleclip,abdal2021styleflow}. However, as these models are domain-specific and trained only on faces, they are limited to editing and generating cropped portrait images. 

This raises the following question - \textit{How can we combine the generalized knowledge from T2I models with the face-specific knowledge from StyleGAN models?} Such a framework will enjoy benefits from both groups, enabling coarse control with text and fine-grained attribute control through latent manipulation in the generation process. In this work, we propose a novel approach to combine these two categories of models by \textit{conditioning the T2I model with $\mathcal{W+}$ space from StyleGAN2}. Conditioning on the $\mathcal{W+}$
provides a natural way for embedding faces in T2I model by projecting them into $\mathcal{W+}$ space using existing StyleGAN2 encoders~\cite{e4e}. 
The design of having $\mathcal{W+}$ as the inversion bottleneck has two major advantages: \textit{1) excellent inversion of a face with precise reconstruction of attributes, and 2) explicit control over facial attributes for fine-grained attribute editing.} 
To the best of our knowledge, this is the first work to demonstrate the combination of two powerful generative models StyleGANs and T2I diffusion models for controlled generation.

To condition the T2I model on the $\mathcal{W+}$ space, we train a latent adaptor - a lightweight MLP, conditioned on the diffusion process's denoising timestep. It takes a latent code $w \in \mathcal{W+}$ of a face as input to predict a pair of time-dependent token embeddings that represent the input face and are used to condition the diffusion model cross-attention. We observe that having a different embedding for each timestep provides more expressivity to the inversion process. The latent adaptor is trained on a dataset of (image, $w$) latent pairs, guided by identity loss, a class regularization loss, and standard denoising loss. To further improve the inversion quality, we perform a few iterations of subject-specific U-Net tuning on the given input image using LoRA~\cite{lora}. The embedded subject can then be edited in two ways: \textit{i) coarse semantic edits using text (for e.g., changing the layout and background) and ii) fine-grained attribute edits by latent manipulation in $\mathcal{W+}$ (for e.g., smooth interpolation through a varying range of smiles, ages)}. Some example edits are provided in Fig.~\ref{fig:teaser}. Our method \textit{generalizes the fine-grained attribute edits from cropped faces (in StyleGANs) to in-the-wild and stylized face images} generated by T2I diffusion model.


The proposed method can be easily extended for multiple-person generation, which requires high fidelity identity of all the subjects (Fig.~\ref{fig:teaser}). We first predict separate token embeddings for each person and then perform subject-specific tuning to obtain personalized models. However, training a single personalized model for multiple subjects results in the problem of \textit{attribute mixing} (Fig.~\ref{fig:multi-person-ablate}) between faces, where attributes from one face are mixed with another. Instead, we learn separate subject-specific LoRA models, which are then jointly inferred with a chained diffusion process. The intermediate outputs for these processes are merged using an instance segmentation mask after each denoising step. This framework resolves \textit{attribute-mixing} among subjects and preserves the identity from the fine-tuned models. The attributes of individual subjects can be edited in a fine-grained manner in $\mathcal{W+}$ while preserving the other subjects' attributes as shown in Fig.~\ref{fig:teaser}-(Attribute Editing).

We perform extensive experiments for embedding single and multiple subjects in StableDiffusion~\cite{ldm} model. Compared to the existing personalization method, the proposed method is extremely efficient and achieves a good tradeoff between identity preservation and text alignment. Next, we present results for fine-grained attribute editing with continuous control in the $\mathcal{W+}$ latent space and compare performance with existing editing methods. Finally, we compare the results for composing multiple subjects and attribute editing of individual subjects. In summary, our primary contributions are as follows:

\begin{itemize}
    \item First approach to combine large text-to-image models with StlyeGAN2 by conditioning T2I on rich $\mathcal{W+}$ latent space 
    \item Effective personalization method using a single portrait image enabling fine-grained attribute editing in $\mathcal{W+}$ space and coarse editing with text prompts. 
    \item Novel approach to fuse multiple personalized models with chained diffusion processes for multi-person composition. 
\end{itemize}

\section{Related Work}
\textbf{Text-based image generation.}
Large text-to-image diffusion models~\cite{imagen,ldm,dalle-2} achieve excellent image generation performance when trained on internet scale captioned image datasets~\cite{schuhmann2022laion}. These models are scaled to high resolution by learning cascaded diffusion models~\cite{dalle,dalle-2,imagen} that generate low-resolution images followed by upsampling. Another promising approach is to train diffusion models in the compressed latent space of a pretrained autoencoder~\cite{ldm}.


\noindent \textbf{Personalization} aims to embed a concept in T2I model, given a few input images. One group of methods optimize for object-specific token embeddings ~\cite{textual-inversion,neti,celeb-basis} via optimization. These approaches preserve text editability, however, they struggle to preserve identity. Another direction is based on fine-tuning diffusion model with strong regularization to avoid overfitting ~\cite{ruiz2023dreambooth, custom-diffusion, imagic}. The third set of methods~\cite{e4t, profusion, ruiz2023hyperdreambooth, xiao2023fastcomposer, ipadaptor, person_new_face0, wang2024instantid,person_new_enhance_det} learns a shared domain-specific encoder for faster inversion by leveraging the class-specific features.


\noindent \textbf{Embedding faces.} Recently embedding human faces in T2I models has received a lot of attention ~\cite{celeb-basis,xiao2023fastcomposer,e4t,chen2023photoverse,profusion} as the generic personalization methods~\cite{textual-inversion,ruiz2023dreambooth} often fail to faithfully embed human faces. Celeb-basis~\cite{celeb-basis} learns a basis of the celebrity names in the token embedding space. The weights of these basis vectors are then predicted by an encoder model applied to the input image. Profusion~\cite{profusion} proposes a regularization-free encoder-based approach. Photoverse~\cite{chen2023photoverse} applies a dual branch conditioning in text and image domains for faster and more accurate inversion of faces. Although these methods are able to achieve good inversion, they do not allow for fine-grained attribute control. A concurrent work ~\cite{sg2diff} aims to map the $\mathcal{W+}$ space to the T2I model, however, their method is limited in preserving identity.



\noindent \textbf{Image editing.} Trained T2I models serve as strong image priors and enable various image editing and restoration applications~\cite{meng2021sdedit, shape-variations, brooks2023instructpix2pix, wang2023exploiting}. For fine-grained image editing ~\cite{prompt2prompt,shape-variations,pix2pix-zero} localizes the object in the image space using attention masks and only allows editing in the specified region.
However, these methods rely on text to change the localized object, which does not allow for fine-grained control. Promising approaches like  ~\cite{brack2023sega,gandikota2023concept} provide a finer control by interpolation in the noise space or training special sliders per attribute. Another set of works explores the intermediate feature space of unconditional diffusion models to obtain finer attribute control in generation~\cite{hspace,parihar2024balancing}. However, they are limited to editing of generated images and not personalized subjects. We take inspiration from GAN models to attain fine-grained attribute control for real subjects by leveraging their disentangled and smooth latent spaces~\cite{interface,hsr}. This enables precise attribute editing through latent manipulation ~\cite{interface,patashnik2021styleclip,abdal2021styleflow,parihar2022everything} and we aim to embed these properties in pretrained T2I models. 


\section{Method} 

\subsection{Preliminaries}
\noindent \textbf{Text-to-image Diffusion Models.}
This work uses StableDiffusion-v2.1~\cite{ldm} as a representative Text-to-image (T2I) diffusion model. Stable diffusion is based on the latent diffusion model, which applies the diffusion process in the latent space. Its training involves two stages: a) training a VAE or VQ-VAE autoencoder to map images to a compressed latent space, and b) training a diffusion model in its latent space conditioned on text for guiding the generation. This framework disentangles the learning of fine-grained details in the autoencoder and semantic features in the diffusion model, resulting in easier scaling. 

\noindent \textbf{Style-based GANs,} ~\cite{sg,sg2,eg3d} have been widely adapted to generate realistic object-specific images such as faces. Further, these models have disentangled latent space, which enables smooth interpolation between images and fine-grained attribute editing~\cite{interface, patashnik2021styleclip}. These properties are induced by mapping the Gaussian latent space to a learned latent space $\mathcal{W}/\mathcal{W}+$ with a mapper network. Further, GAN encoder models~\cite{e4e,psp} can encode and edit real images that invert a given image into $\mathcal{W+}$ space, allowing for fine-grained editing of real images. 


\begin{wrapfigure}{r}{0.5\textwidth}
    \centering
    \vspace{-15mm}
    \includegraphics[width=1.0\linewidth]{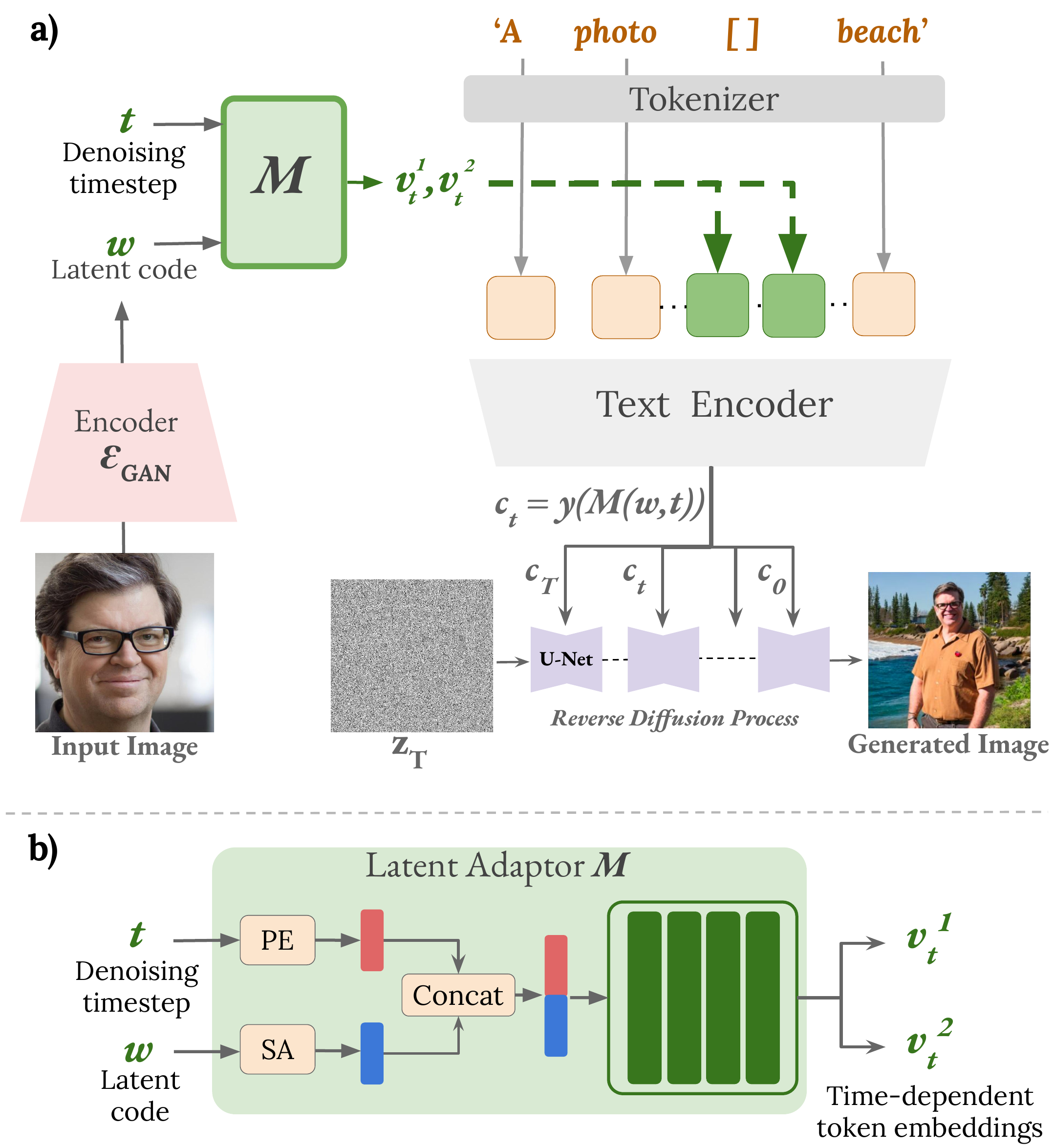}
    \caption{\textbf{Framework for personalization.} Given a single portrait image, we extract its $w$ latent representation from encoder $\mathcal{E}_{GAN}$. The latent $w$ along with diffusion timestep $t$ are passed through the latent adaptor $\mathcal{M}$ to generate a pair of time-dependent token embeddings $(v_t^1, v_t^2)$ representing the input subject. Finally, the token embeddings are combined with arbitrary prompts to generate customized images.}
    \vspace{-7mm}
    \label{fig:method-arch}
\end{wrapfigure}

\vspace{-4mm}
\subsection{Overview}
While the T2I models trade off the diversity in generation with an attribute-rich latent space, our goal is to condition the T2I model with an attribute-rich - $\mathcal{W+}$ space from StyleGAN2, that allows for disentangled and fine-grained control over the face attributes in the generated image. To condition the T2I model on $\mathcal{W+}$, we augment the T2I model with a learnable latent adaptor network $\mathcal{M}$ that projects a latent code $w \in \mathcal{W+}$ into the text embedding space. For embedding a new subject, we pass it through a pre-trained StyleGAN2 encoder $\mathcal{E}_{GAN}$\cite{e4e} to obtain $w$ latent code, which is passed through $\mathcal{M}$ to obtain the corresponding text-embedding as shown in Fig.~\ref{fig:method-arch}a). Conditioning on $\mathcal{W+}$ enables fine-grained attribute control in the generated image by latent manipulation. In the next sections, we discuss the details of the proposed latent adaptor, model training, and fine-grained attribute editing.



\vspace{-5mm}
\subsection{Latent adaptor $\mathcal{M}$} 
\vspace{-1mm}
We implement the latent adaptor $\mathcal{M}$ as a shallow MLP network that maps the $w$ latent code from StyleGAN to the token embedding space of the T2I model for any human face image as input. We learn two token embeddings $(v^1, v^2)$ to represent a human subject as it is known to improve the embedding quality ~\cite{celeb-basis}. To extract the timestep-specific semantic information from the latent $\mathbf{w}$, we condition $\mathcal{M}$ on the diffusion timestep $t$, as diffusion models represent the semantic hierarchy in a timestep-wise fashion~\cite{patashnik2023localizing}. The output of $\mathcal{M}$ is a set of pair of embedding vectors $\{(v_t^1, v_t^2)\}_{t=0}^{t=T}$, a pair for each timestep $t$. Time-dependent token embeddings allow for a richer representation space and improve identity preservation (shown in Fig.~\ref{fig:quant-results-embed}). The complete architecture of $\mathcal{M}$ is shown in Fig.\ref{fig:method-arch}b). The input $t$ is first passed through positional encodings~\cite{tancik2020fourier} and the flattened $w$ latent code is passed through a self-attention layer to get relevant features. The encoded representations are then concatenated before passing through a set of linear layers. The obtained pair embedding $(v_t^1, v_t^2)$ represents the person and is then passed at $t^{th}$ denoising time-step in the U-Net for generation.

\subsection{Training} 
We perform a two-stage training, where we first pretrain latent adaptor $\mathcal{M}$ on a face dataset, followed by an few iterations of subject-specific training of $\mathcal{M}$ and diffusion U-Net with low-rank updates for improving identity as detailed below.


\noindent \textbf{Pretraining.} The mapper $\mathcal{M}$ is pretrained with a paired dataset $\mathcal{D}_w$ consisting of $(I, w)$ pairs, where $I$ is a portrait face image and $w$ is its corresponding latent code obtained as $\mathcal{E}_{GAN}(I)$. During training, we sample a pair $(I, w)$ and a denoising timestep $t \in (1,T)$ which are passed through $\mathcal{M}$ to obtain the pair of token embeddings $(v_t^1, v_t^2)$ corresponding to the input subject. We place the sampled tokens along with the neutral prompt - $y=$\textit{`A photo of a ... person'} and pass through the text encoder to obtain the final text embeddings $c(y(\mathcal{M}(t,w)))$. We add the noise from the noise schedule at $t$ to the image $I$ and train diffusion loss along with additional regularization losses shown in Eq.\ref{eq:eq1} to train $\mathcal{M}$. In this stage of training, all the modules - $\mathcal{E}_{GAN}$, text-encoder, and U-Net are frozen, except for $\mathcal{M}$ which is a shallow MLP, making the training compute efficient. 


\begin{wrapfigure}{r}{0.6\textwidth}
    \centering
    \vspace{-7mm}
    \includegraphics[width=1.0\linewidth]{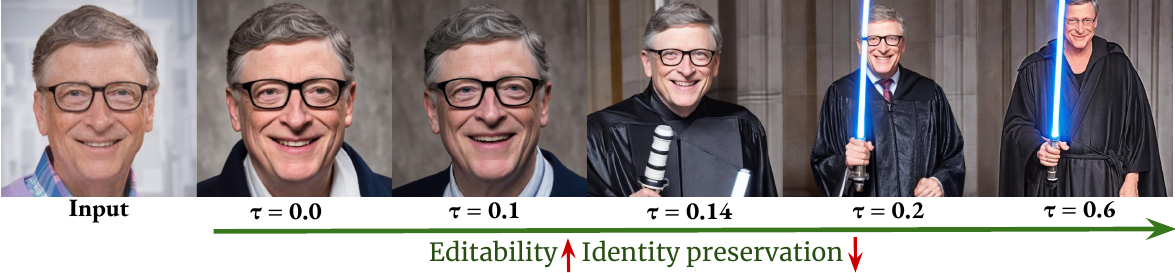}
    \vspace{-6mm}
    \caption{Delayed identity injection results in better text editability.}
    \vspace{-9mm}
    \label{fig:delay-injection}
\end{wrapfigure}

\noindent \textbf{Subject specific training.} In the second stage, we fine-tune encoder $\mathcal{M}$ and U-Net for a few iterations with the single input image. Specifically, we perform low-rank weight updates (LoRA~\cite{lora}) on the U-Net projection matrices and fine-tune $\mathcal{M}$, using the combined loss from Eq.~\ref{eq:eq1}. In LoRA training, the model weights are updated as $\mathbf{W} = \mathbf{W} + \alpha \mathbf{\Delta} W$, where $\mathbf{\Delta} W$ is the learned low-rank residual weights. The hyper-parameter $\alpha$ controls the extent of fine-tuning and allows for a trade-off between identity preservation and text editability.
This second stage of low-rank tuning improves the subjects' identity without hurting the text editability (Fig.~\ref{fig:quant-results-embed}). 

\noindent \textbf{Loss function.}
We train latent adaptor with a combination of denoising diffusion loss $\mathcal{L}_{Diffusion}$ and regularization loss $\mathcal{L}_{reg}$ following ~\cite{e4t}. The diffusion loss enforces text-to-image consistency, and regularization loss ensures that the predicted token embedding is close to the token embedding of the superclass $v_{cls}$, such as $\textit{face}$. Additionally, we add identity loss $\mathcal{L}_{ID}$ defined as the MSE between the face recognition embeddings from ~\cite{curr-face-rec} to preserve the identity during inversion. The final loss is computed as a linear combination of these losses: 

\vspace{-5mm} 
\begin{equation}
\label{eq:eq1}
\begin{split}
    \mathcal{L}_{Diffusion} &= E_{z,y,\epsilon,t} [||\epsilon - \epsilon_{\theta}(z,c(y(\mathcal{M}(t,w)))||_2^2] \\ 
     \mathcal{L}_{reg} &=  ||\mathcal{M}(t,w) - v_{cls}||_2^2 \\ 
     \mathcal{L}_{ID} &= ||\mathcal{E}_{ID}(x_t) - \mathcal{E}_{ID}(I)||_2^2 \\  
     \mathcal{L} &= \mathcal{L}_{Diffusion} + \lambda_{reg}\mathcal{L}_{reg} + \lambda_{ID}\mathcal{L}_{ID} 
\end{split}
\end{equation}
\vspace{-3mm}

\noindent where $\lambda_{reg}$ and $\lambda_{ID}$ are hyper-parameters and $\mathcal{E}_{ID}$ is pretrained face-recognition model~\cite{curr-face-rec}. To compute $\mathcal{L}_{id}$ at the intermediate denoising step, we use DDIM~\cite{ddim} approximation of the clean image $\hat{x_0}$ and pass it to the face detector. 


\subsection{Inference}

\begin{wrapfigure}{r}{0.5\textwidth}
    \vspace{-8mm}
    \includegraphics[width=1.0\linewidth,]{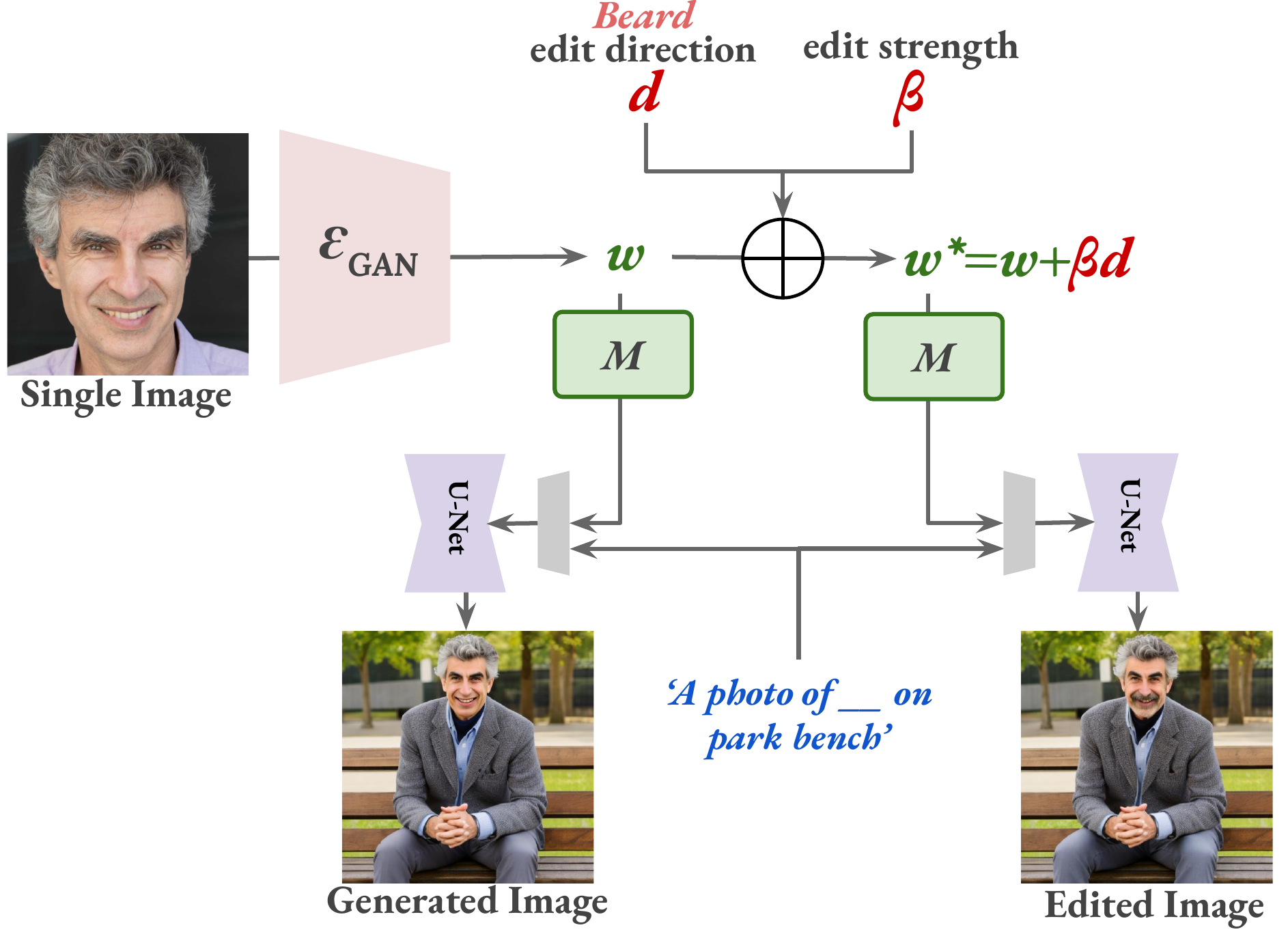}
    \vspace{-4mm} 
    \caption{\textbf{Fine-grained attribute editing.} We map the given input image into $w$ latent code, which is shifted by a global linear attribute edit direction to obtain edited latent code $w*$. The edited latent code $w*$ is then passed through the T2I model to obtain fine-grained attribute edits. The scalar edit strength parameter \textcolor{red}{$\beta$} can be changed to obtain continuous attribute control.} 
    \vspace{-7mm}
    \label{fig:attr-edit}
\end{wrapfigure} 

During inference, given a single image $I$, we obtain its token embedding as $(v_t^1, v_t^2) =\mathcal{M}(\mathcal{E}_{GAN}(I), t)$ for all the time steps $t \in (1,T)$. These embeddings can be added with text prompts to generate a novel composition of the learned subject. The image generation process in diffusion models follows a hierarchical structure, where the layout is formed in the first few steps, followed by the formation of object shape and appearance ~\cite{shape-variations}. As our primary aim is to embed a subject's identity, we inject the obtained token embedding only after a time threshold ($t < \tau$) to not hurt the layout generated during the initial timesteps.


\noindent For the initial denoising timesteps ($t > \tau$), we use a celebrity name as a placeholder in the prompt, e.g.\, \textit{`A photo of \underline{Brad Pitt} as a star wars character'} as the model generates improved image layouts when prompted popular subjects. Empirically, we observe that the generations are not sensitive to the text celebrity name used, and it acts as a placeholder, so we fix a single celebrity name, and there is no overlap of the identity with the dataset used for evaluation. This delayed injection of the learned embedding improves text alignment, and passing the predicted token embeddings to all the timesteps results in poor compositions where the model output is a cropped face, as shown below in Fig.~\ref{fig:delay-injection}. 

\subsection{Fine-grained control over face attributes}
\label{subsec:fgcontrol}
Once trained, the latent adaptor $\mathcal{M}$ bridges between disentangled and smooth $\mathcal{W}+$ latent space and the text-conditioning of the diffusion model. This enables the \textit{transfer} of latent attribute editing methods that function in the $\mathcal{W+}$ space of StyleGANs~\cite{interface,patashnik2021styleclip,harkonen2020ganspace} to the diffusion model. Specifically, for a given source image $I_s$, we first obtain its corresponding $w$ latent code with $\mathcal{E}_{GAN}$. Next, we edit the latent code $w$ by adding a \textit{global} linear attribute edit direction $d$ with scalar weight $\beta$ to obtain $\hat{w} = w + \beta d$. Note, the same global edit direction $d$ generalizes for all the identities in the $\mathcal{W}+$ space~\cite{interface}. The edited latent code $\hat{w}$ 

\begin{wrapfigure}{r}{0.5\textwidth}
    \centering
    \vspace{-8mm}
    \includegraphics[width=1.0\linewidth]{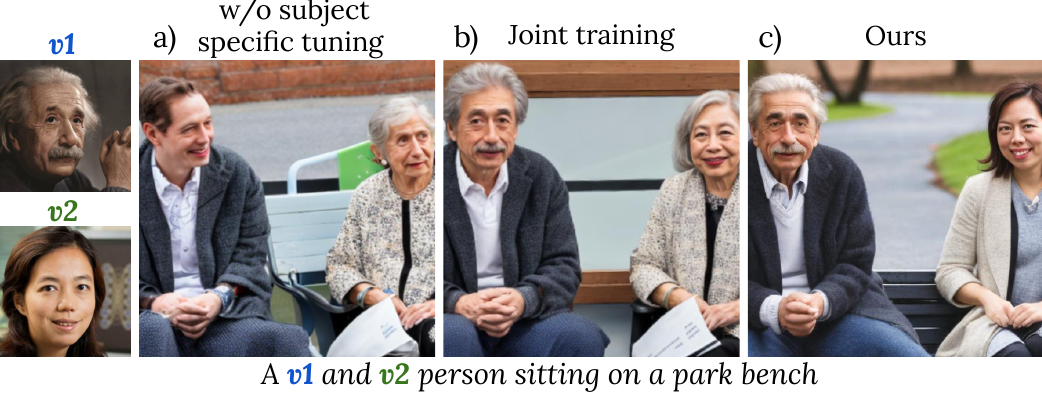}2
    \vspace{-8mm}
    \caption{Composing multiple persons without finetuning results in identity distortion. Finetuning a single model for both the identities results in \textit{attribute mixing}, the age and facial hairs from \textbf{v1} are transferred to \textbf{v2}. Combining outputs of individual finetuned models results in excellent identity preservation without \textit{attribute mixing}.}
    \vspace{-8mm}
    \label{fig:multi-person-ablate} 
\end{wrapfigure}  

\noindent 
is then passed through $\mathcal{M}$ to obtain the edited token embedding $\hat{v}_t$. Notably, one can precisely control the strength of the attribute edit by changing the scalar $\beta$ as shown in Fig.~\ref{fig:fine-grained-single-edit}. To preserve the scene layout during editing, we use the same starting noise and copy the self-attention maps obtained during the generation with unedited $w$ similar to~\cite{shape-variations}. Further, one can easily combine multiple edit directions by taking a weighted combination of individual attribute edits (Fig.~\ref{fig:fine-grained-multi-edit}), thanks to the linearity of the $\mathcal{W+}$. 

\subsection{Composing multiple persons} 
\label{sec:multi-person} 

\begin{wrapfigure}{r}{0.5\textwidth}
    \vspace{-10mm}
    \includegraphics[width=1.0\linewidth]{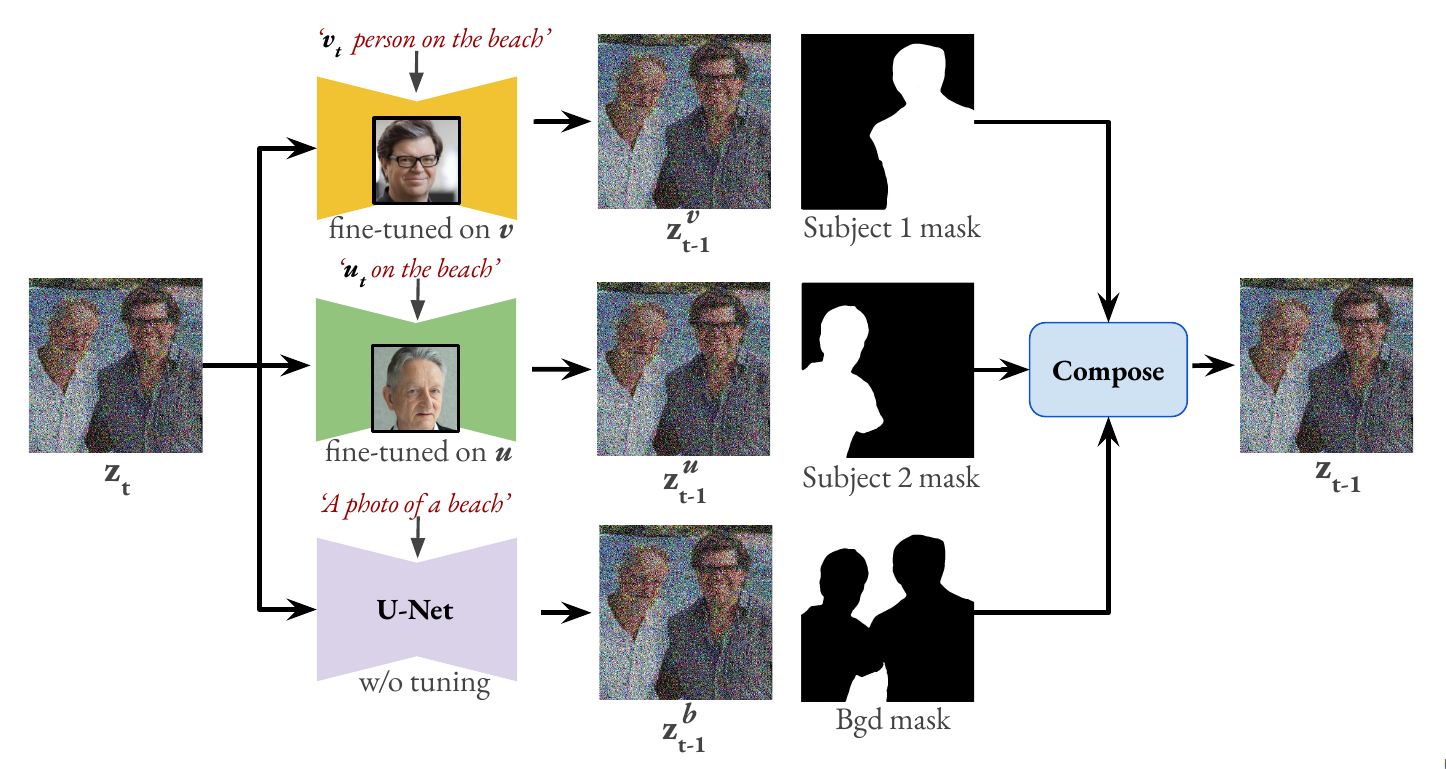}
    \vspace{-8mm}
    \caption{\textbf{Composing multiple subjects.} We run multiple parallel diffusion processes, one per subject and one for the background, which are fused using instance masks at each denoising step. Importantly, the diffusion process for each subject is passed through its corresponding fine-tuned model, which results in excellent identity preservation.} 
    \vspace{-10mm}
    \label{fig:multi-person-composition}
\end{wrapfigure}  

Our method can be extended to compose multiple subject identities in a single scene. Naively, embedding multiple token embeddings (one per subject) in the text prompt without subject-specific tuning results in identity distortion Fig.~\ref{fig:multi-person-ablate}a). Jointly performing subject-specific tuning improves the identity but suffers from \textit{attribute mixing}, where facial attributes from one subject are transferred to another, such as age and hairs in Fig.~\ref{fig:multi-person-ablate}b). This is a well-known issue in T2I generation, where the model struggles with multiple objects in a scene and binds incorrect attributes~\cite{attend-excite}. We take an alternate approach inspired by MultiDiffusion~\cite{bar2023multidiffusion}, where we run multiple chained diffusion processes, one for each subject and one for the background. The outputs of these processes are combined at each denoising step using an instance segmentation mask. We run the diffusion process for each subject through its corresponding subject-specific finetuned model. This preserves the subjects' details learned by each finetuned model and enables high-fidelity composition of multiple persons without \textit{attribute mixing}. To obtain an instance segmentation mask, we run a single diffusion process with a prompt containing two persons and apply the off-the-shelf segmentation model SAM~\cite{sam} on the generated image. Further, we can perform fine-grained attribute edits on a single subject with latent manipulation in $\mathcal{W+}$ space while preserving other subjects, as shown in Fig.~\ref{fig:teaser}.

\begin{figure*}[t]
    \centering
\includegraphics[width=1.0\linewidth]{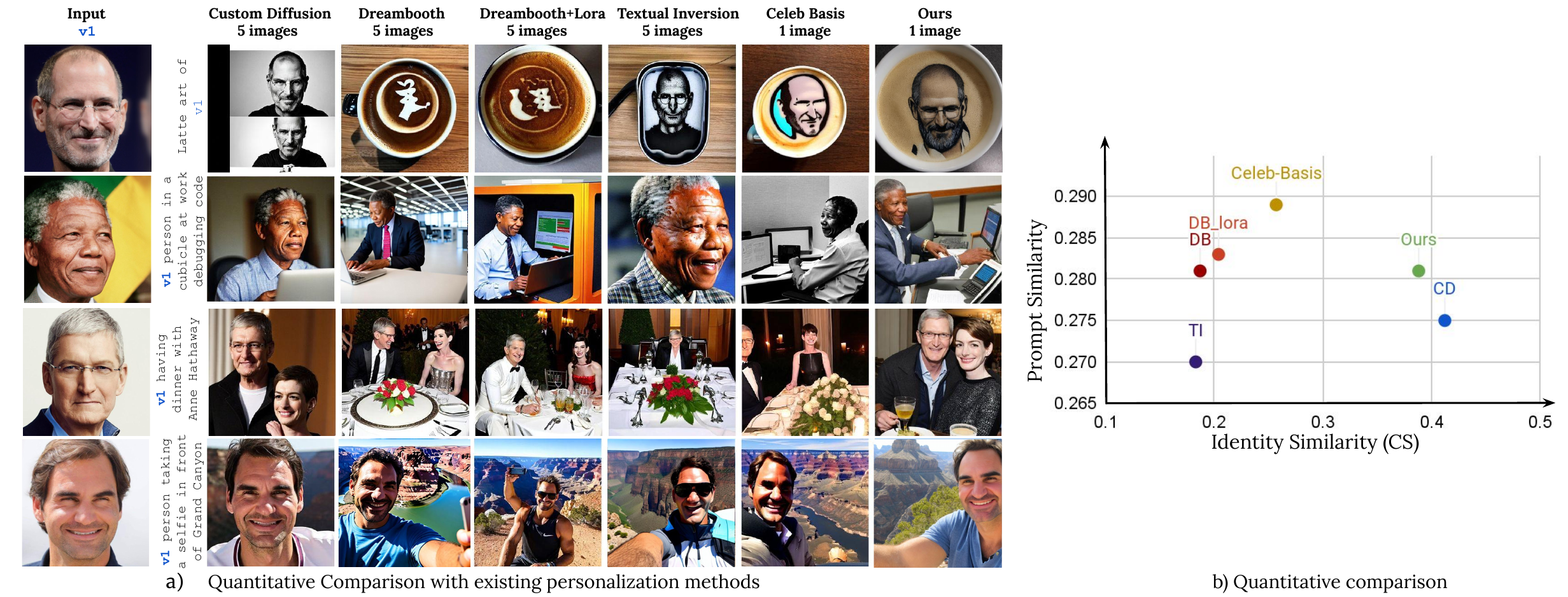}
    \vspace{-6mm}
    \caption{
    \textbf{Comparison for single subject personalization.} Existing personalization methods designed for generic concepts either achieve good Identity similarity (Custom Diffusion) or Prompt similarity (Celeb Basis, Dreambooth, Dreambooth+Lora) but fail to achieve both simultaneously. Like ours, Celeb Basis is a single image, face-specific personalization method that achieves good prompt similarity. However, faces generated by Celeb Basis have a cartoonish look and lack realism. Our method strikes a perfect balance between Identity similarity and Prompt similarity, as shown in the plot, and generates highly photo realistic images following the text.} 
    \vspace{-4mm}
    \label{fig:qual_compare_single_attribute} 
\end{figure*}

\section{Experiments} 
\label{sec:experiments}
We perform all our experiments on StableDiffusion-v2.1~\cite{ldm} as a representative T2I model. For inversion, we use pre-trained StyleGAN2 e4e encoder~\cite{e4e} trained on the face dataset to map images in $\mathcal{W}+$. In the following sections, we first discuss the datasets and metrics (Sec.\ref{subsec:data_metric}), followed by results on single-subject and multi-subject personalization (Sec.\ref{subsec:single-person-res}), fine-grained attribute editing (Sec.\ref{subsec:fine-grained-editing}), and ablation studies (Sec.\ref{subsec:ablations}). 

\subsection{Dataset and metrics}
\label{subsec:data_metric}
\noindent \textbf{Dataset.} The latent adaptor is trained with a combination of synthetic images generated from StyleGAN2 and real images from the FFHQ~\cite{sg} dataset. The dataset contained $70K$ image and corresponding $w$ latent codes obtained from e4e~\cite{e4e}. We collected a dataset of $30$ subjects for evaluation, including scientists, celebrities, sports persons, and tech executives. We also evaluate on 'non-famous' identities and synthetic faces in supplementary. We use a set of $25$ diverse text prompts, including texts for stylization, background change, and doing certain actions. Further details about the setup is provided in the supplementary.

\noindent \textbf{Metrics.} We evaluate the personalization performance using two widely used metrics for subject personalization: \textbf{Prompt similarity} - to measure the alignment of the prompt with the generated image using CLIP~\cite{clip}, and \textbf{Identity similarity}(\textbf{CS}) - to measure the identity similarity between the input image and the generated image using cosine similarity between face embeddings from~\cite{wang2018cosface}. To evaluate fine-grained attribute editing, we compute the change in Prompt similarity \textbf{($\Delta$ CLIP)} with the attribute prompt (e.g., `A \textit{v1} person smiling') before and after the edit. Additionally, we measure the change in the image during editing with \textbf{LPIPS}~\cite{zhang2018unreasonable} and Identity similarity. For an ideal fine-grained attribute edit, a higher $\Delta$ CLIP indicates a meaningful edit and a lower LPIPS and higher ID-sim denotes the preservation of source identity. 

\begin{wrapfigure}{r}{0.5\textwidth}
    \vspace{-8mm} 
    \includegraphics[width=1.0\linewidth]{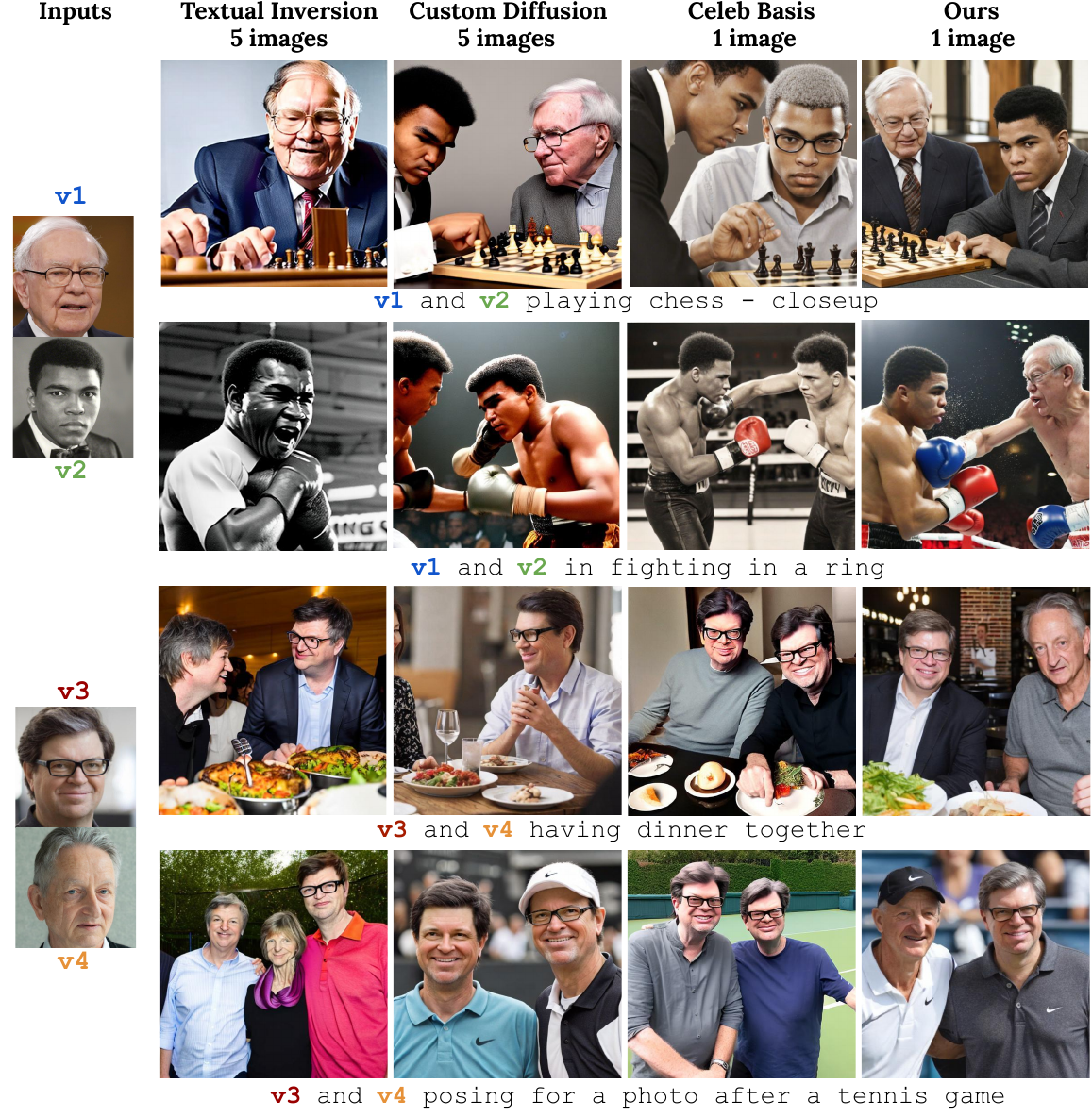}
    \vspace{-8mm}
    \caption{\textbf{Comparison for multi-subject generation.} Textual Inversion struggles to generate both the subjects and distorts their identities. Custom Diffusion and Celeb Basis can compose but suffer from \textit{attribute mixing} - the same hairstyle and facial features are copied to both faces. This effect is more pronounced in Celeb Basis. Our method disentangles the attributes of the two subjects and generates identity-preserving composition.}
    \label{fig:multi-person}
    \vspace{-6mm}
\end{wrapfigure}

\subsection{Comparison with personalization methods.}
\label{subsec:single-person-res}
\begin{figure*}[]
    \centering
    \vspace{-6mm}
    \includegraphics[width=0.90\textwidth]{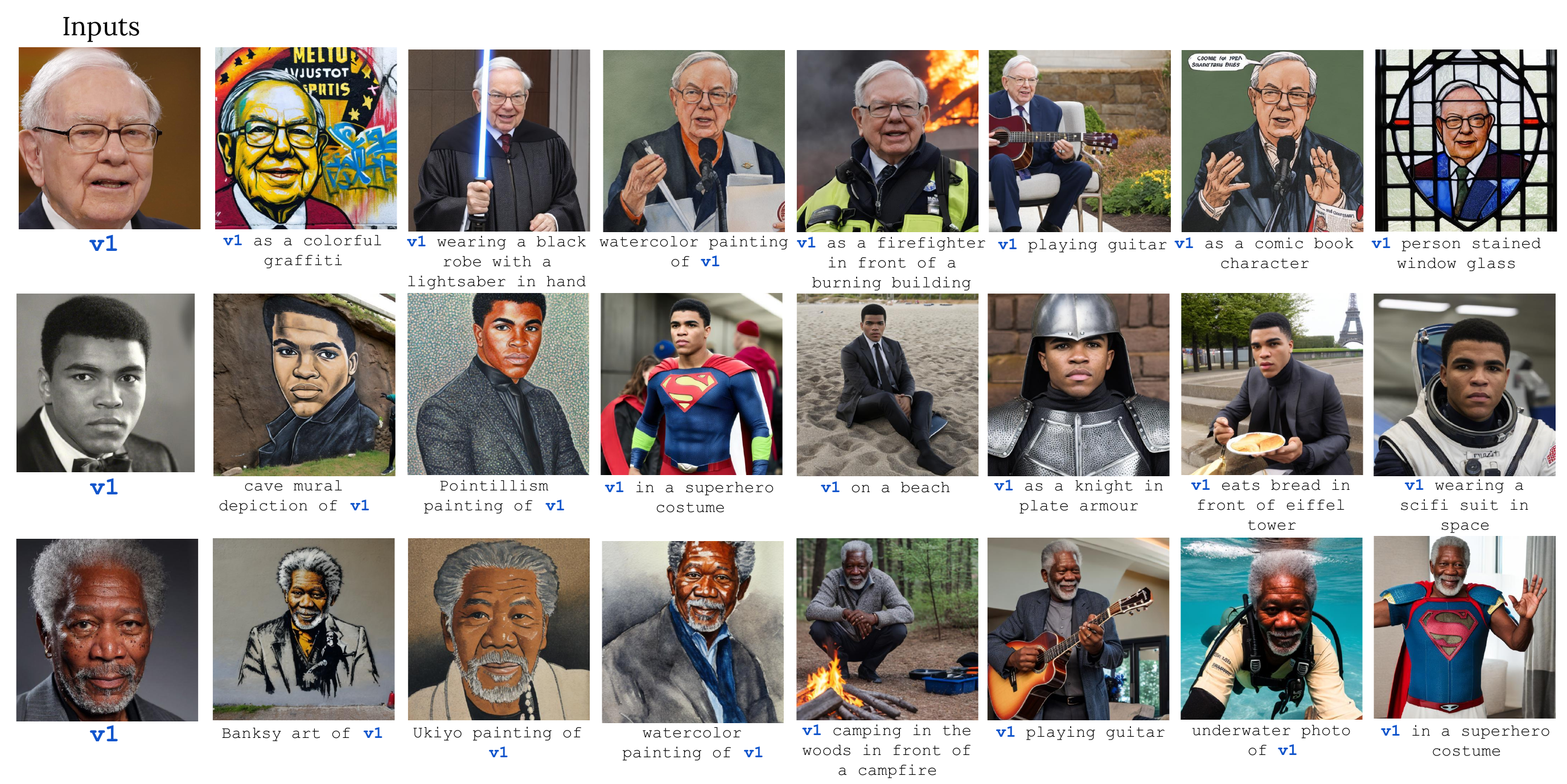}
    \vspace{-2mm}
    \caption{\textbf{Additional results for single subject personalization.} Our method achieves excellent realism and text alignment with the prompts.}
    \vspace{-6mm} 
    \label{fig:extra-single-subject-personalization-a)}
\end{figure*}

\textbf{Single-subject personalization.} We perform single-image personalization on the evaluation set with diverse text prompts in Fig.~\ref{fig:qual_compare_single_attribute}, ~\ref{fig:extra-single-subject-personalization-a)} \& S8. We compare with following fine-tuning-based personalization methods: Custom Diffusion~\cite{custom-diffusion}, Dreambooth~\cite{ruiz2023dreambooth}, Dreambooth+LoRA, which is Dreambooth with low-rank updates to avoid overfitting, Textual Inversion~\cite{textual-inversion} and Celeb Basis~\cite{celeb-basis}. All the methods are trained with $5$ images per subject except Celeb-basis and ours, which operate on a single input image. Details about hyper-parameters for competitor methods are provided in the supplementary. Custom Diffusion embeds a subject while preserving its identity; however, it mostly generates closeup faces and does not follow the text prompts to stylize the subject or to have it perform an action. Dreambooth cannot embed the subject's identity faithfully, whereas with LoRA training, the identity is improved along with text alignment, which helps avoid overfitting. Textual Inversion and Celeb Basis have poor identity preservation as they fine-tune only the token embedding and not the U-Net. Celeb Basis achieves the highest text alignment due to the strong regularization imposed by basis spanning across celebrity names. Our method strikes a perfect balance between text alignment and identity preservation. Note that ours and the Celeb Basis use only $1$ input image, which slightly affects the identity, against Custom-diffusion that requires $5$ images. We have provided an additional comparison with encoder-based models and the recent IP-adaptor~\cite{ipadaptor} method in supplementary material. 

\begin{wrapfigure}{r}{0.5\textwidth}
    \vspace{-10mm}
    \includegraphics[width=1.0\linewidth]{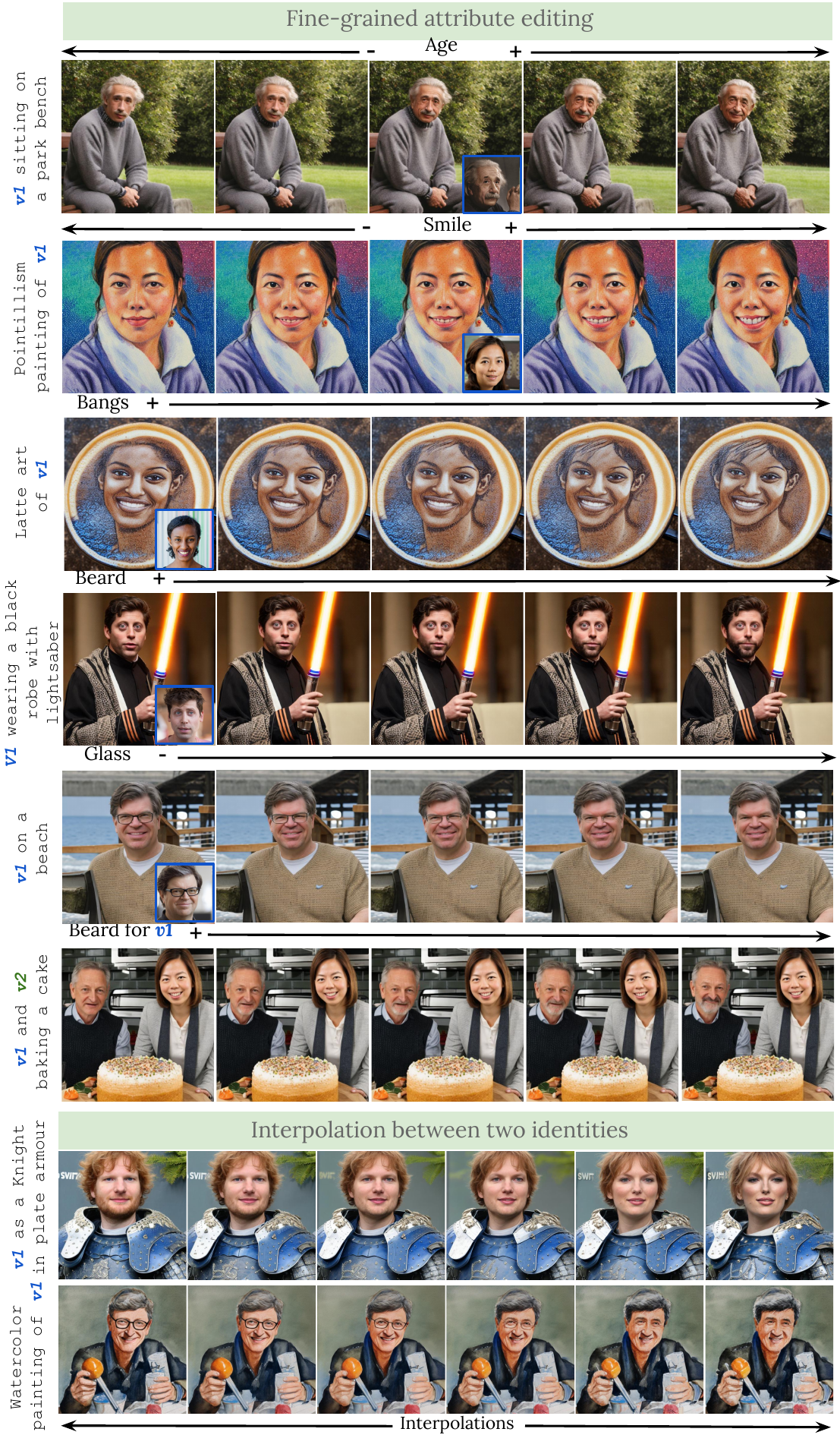}
    \vspace{-6mm}
    \caption{\textbf{Attribute Control.} We perform continuous attribute editing by adding an attribute edit direction in $\mathcal{W+}$ and increasing its edit strength $\beta$. Our method performs disentangled edits for various attributes while preserving identity and generalizing to in-the-wild faces, styles, and multiple persons. \textbf{Identity Interpolation.} We can perform smooth interpolation between identities by interpolating between the corresponding $w$ codes.}
    \label{fig:fine-grained-single-edit}
    \vspace{-18mm}
\end{wrapfigure}  

\begin{figure*}[]
    \centering
    \vspace{-6mm} 
    \includegraphics[width=0.90\linewidth]{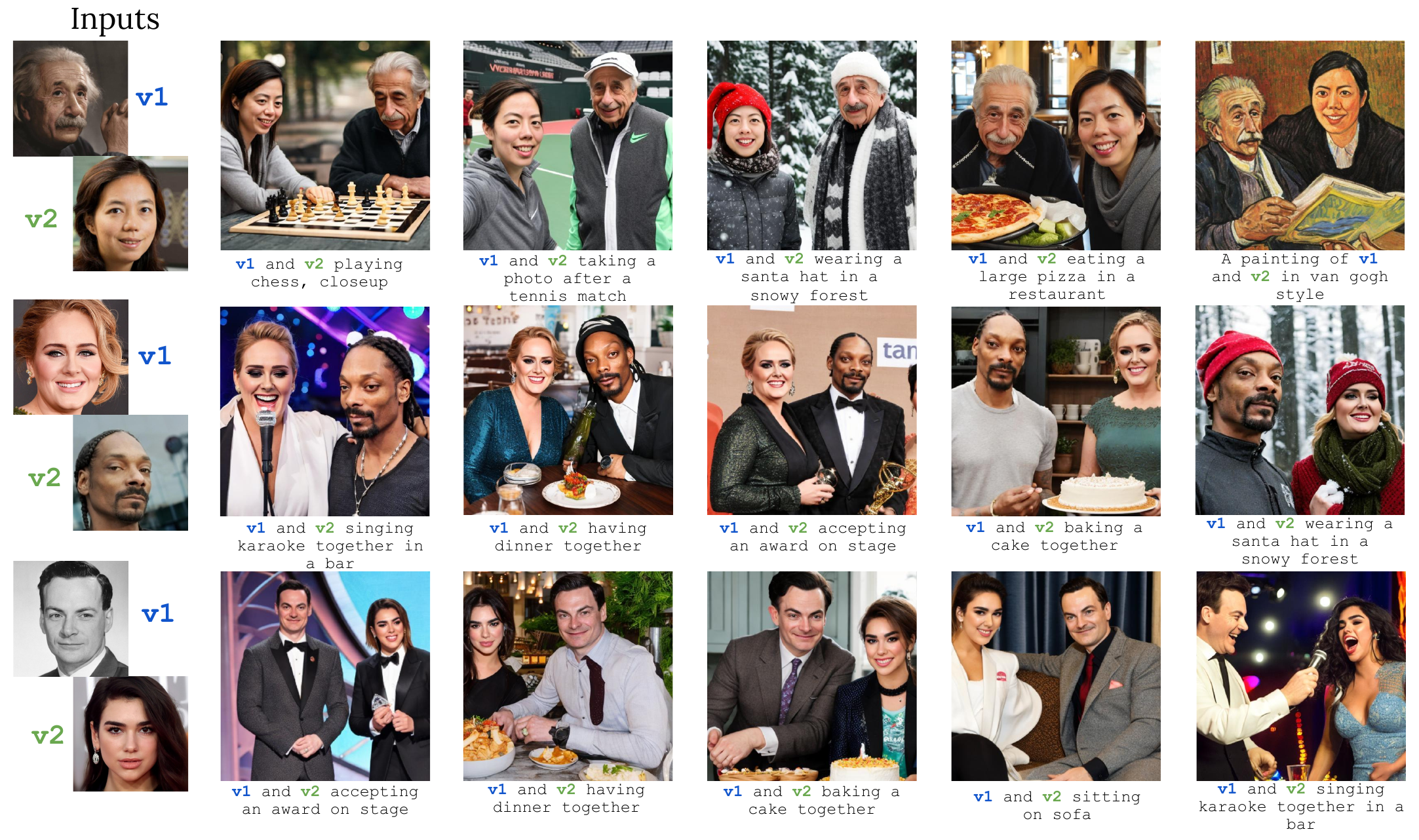}
    \vspace{-2mm}
    \caption{\textbf{Additional results for Multi-subject personalization.} Our method preserves subject identity and follows the text prompt during generation.}
    \label{fig:extra-multiple-subject-personalization} 
\end{figure*}

\noindent \textbf{Multi-subject personalization.} We present results for embedding multiple-person composition in Fig.~\ref{fig:multi-person},~\ref{fig:extra-multiple-subject-personalization}, S7. Specifically, we combine the intermediate outputs of subject-specific tuned models during generation. We compare against multi-concept personalization methods, Textual Inversion, Custom Diffusion, and Celeb Basis. For Textual Inversion and Celeb Basis, we learned two separate token embeddings, one for each subject. For Custom Diffusion, we jointly fine-tune the projection matrices on both subjects. Textual Inversion fails to generate both the subjects in the scene. Celeb Basis and Custom Diffusion generate both the subjects but suffer from \textit{attribute mixing} (eyeglasses from \textit{v4} are transferred to \textit{v3}). As noted earlier, Celeb Basis generates cartoonish faces in most cases. Our method resolves the attribute mixing by running multiple subject-specific diffusion processes and results in highly realistic compositions.

\subsection{Fine-grained control by latent manipulation} 
\label{subsec:fine-grained-editing}
The proposed method matches the disentangled $\mathcal{W+}$ latent space of StyleGANs to the token embedding space of T2I models, allowing for continuous control over the image attributes by latent space manipulation. We present two important image editing applications fueled by the disentangled latent space of StyleGANs: 1) fine-grained attribute editing and 2) smooth identity interpolation. Additionally, our model can restore corrupted face images such as low resolution or inpainting masked facial features in supplementary.

\noindent \textbf{Fine-grained attribute editing.} We perform attribute editing by adding a global latent edit direction in $\mathcal{W+}$ to $w$ encoding of the input image. To have a unified method for all the attributes, we take a simplified approach to obtain edit directions, gathering a small set ($<20$) of paired portrait images before 

\begin{wrapfigure}{r}{0.5\textwidth}
    \vspace{-8mm}
    \includegraphics[width=\linewidth]{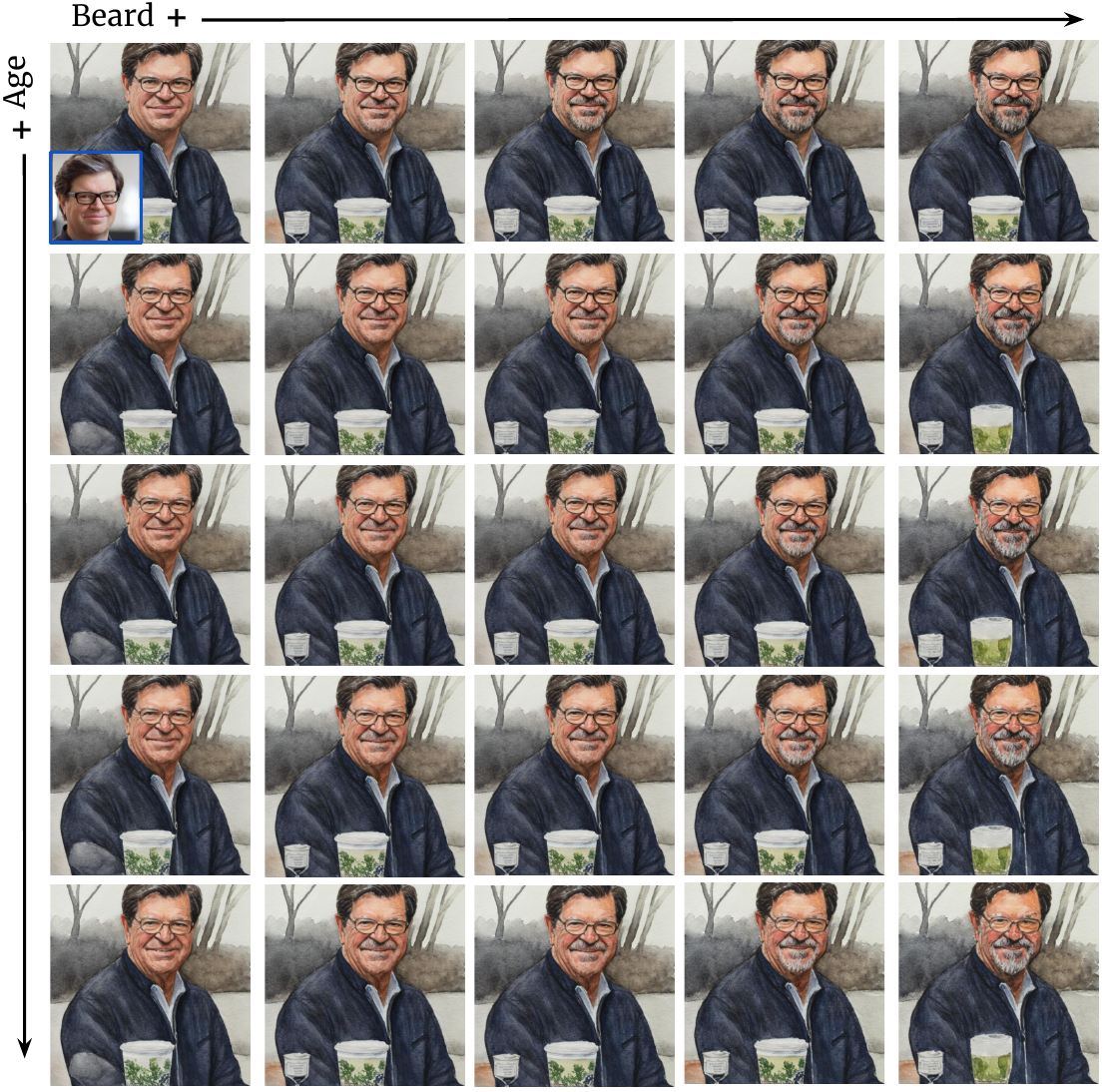}
    \vspace{-6mm}
    \caption{\textbf{Multi-attribute-control.} We can perform continuous edits for two attributes simultaneously by taking a linear combination of attribute edit directions. Observe the smooth and disentangled edit transformations for age and beard attributes while preserving identity.}
    \vspace{-8mm}
    \label{fig:fine-grained-multi-edit}
\end{wrapfigure}

\begin{wrapfigure}{r}{0.5\textwidth}
    \vspace{-4mm}
    \includegraphics[width=1.0\linewidth]{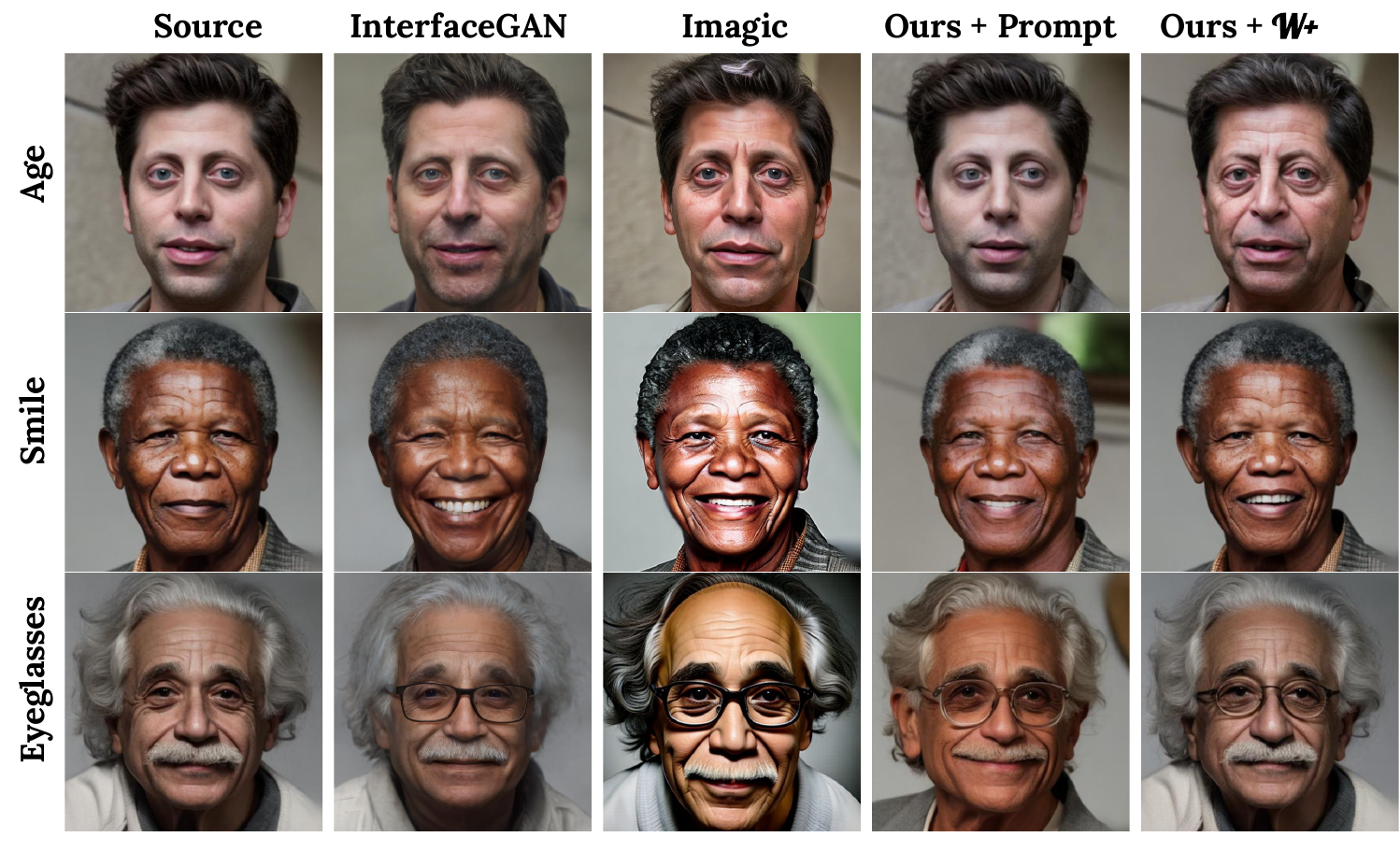}
    \vspace{-8mm}
    \caption{\textbf{Comparison for attribute editing.} Using images for text-based attribute edits results in identity distortion and lacks realism after edit. As both InterfaceGAN and our method leverage the same disentangled latent space, they generate high-quality edits. However, InterfaceGAN is limited to cropped faces while we can edit in-the-wild images too.}
    \label{fig:attr-edit-compare}
    \vspace{-8mm}
\end{wrapfigure}

\begin{table}[t]
\centering
\caption{Comparison of fine-grained attribute editing}
\vspace{-4mm}
\begin{adjustbox}{width=0.9\linewidth} 
\begin{tabular}{@{}c|ccc|ccc|ccc|ccc@{}}
\toprule
               & \multicolumn{3}{c|}{Imagic} & \multicolumn{3}{c|}{InterfaceGAN} & \multicolumn{3}{c|}{Ours + Prompt} & \multicolumn{3}{c}{Ours + $\mathcal{W+}$} \\ \midrule
Attribute      & $\Delta$ CLIP $\uparrow$  & LPIPS $\downarrow$  & CS $\uparrow$  & $\Delta$ CLIP $\uparrow$ & LPIPS $\downarrow$  & CS $\uparrow$  & $\Delta$ CLIP $\uparrow$ & LPIPS $\downarrow$ & CS $\uparrow$  & $\Delta$ CLIP $\uparrow$ & LPIPS $\downarrow$  & CS $\uparrow$ \\ \midrule
\textbf{Beard} & 4.184           & 0.533     & 0.492    & 2.644      & 0.221 & 0.732    & 0.986      & 0.232 &0.877     & 2.473      & 0.185    & 0.731\\
\textbf{Age}   & 4.432           & 0.571     & 0.474    & 1.359      & 0.220 & 0.744    & 0.429      & 0.202 & 0.929    & 1.777      & 0.209    & 0.698\\
\textbf{Smile} & 1.945           & 0.499     & 0.666 & 1.800      & 0.188   & 0.804    & 0.784      & 0.317     & 0.813 & 1.104      & 0.190    & 0.779\\
\textbf{Asian} & 5.908           & 0.585    &0.478  & 4.198      & 0.139    &0.708  & 1.008      & 0.330    & 0.804 & 4.917      & 0.165    & 0.625\\
\textbf{Black} & 5.806           & 0.546    &0.347   & 1.748      & 0.147   &0.825  & 0.897      & 0.222    &0.898  & 1.877      & 0.133    &0.834\\ \bottomrule
\end{tabular}
\end{adjustbox} 
\label{tab:edit-compare} 
\vspace{-6mm} 
\end{table}  

\noindent and after the attribute edit (generated using an off-the-shelf attribute editing method). Next, we take a difference between the corresponding paired $w$ latent and average them to obtain a global edit direction. We found global edit direction for smile, age, beard, gender, race, and eyeglasses. We also show edits with directions obtained using InterfaceGAN~\cite{interface} in Fig. S6 in supplementary. The results for fine-grained control editing are provided in Fig.~\ref{fig:fine-grained-single-edit} \& ~\ref{fig:fine-grained-multi-edit}, where we show disentangled continuous control for various attributes by changing $\beta$ (ref. Sec. \ref{subsec:fgcontrol}) while preserving the identity. Our method \textit{generalizes the edit directions in $\mathcal{W+}$, originally defined for portrait faces, to in-the-wild and stylized face images}. We evaluate attribute editing performance against \textbf{1)} StyleGAN-based global editing method InterfaceGAN ~\cite{interface}, after encoding the image using e4e, \textbf{2)} Prompt-based editing of the learned subject (by giving prompts like `A photo of \textit{v1} smiling'), \textbf{3)} Text-based editing method Imagic~\cite{imagic} build on single image personalization. The quantitative results are present in Tab.~\ref{tab:edit-compare}, and qualitative results are in Fig.~\ref{fig:attr-edit-compare}. Our method achieves the lowest LPIPS scores with high $\Delta$ CLIP, indicating highly disentangled attribute editing. Both text-based editing methods fail to preserve the image regions (higher LPIPS). We achieve high CS scores during edits with higher $\Delta$ CLIP, indicating identity-preserving attribute edits. Ours prompt-based editing achieves a superior CS because the edit is not performed in many cases indicated by lower LPIPS~\cite{zhang2018unreasonable}. Like ours, InterfaceGAN works in $\mathcal{W+}$ latent space and performs similarly in preserving

\noindent the image content, the identity of the subject, and editability. However, it is limited to the editing of portrait faces generated by StyleGANs and loses fine-facial features, whereas our method combines the best of both worlds, allowing for fine-grained latent editing with semantic editing in T2I models.   

\noindent \textbf{Identity interpolation.} $\mathcal{W+}$ space also allows for smooth interpolation between two identities. Given two input images, we obtain their corresponding $w$ latent codes and perform linear interpolation to obtain the intermediate latent codes. When 
\noindent used as conditioning through the latent adaptor, these latents result in realistic face interpolations with smooth changes between the two faces, preserving background, as shown in Fig.~\ref{fig:fine-grained-single-edit}-Bottom.  

\begin{wrapfigure}{r}{0.5\textwidth}
    \vspace{-12mm}
    \includegraphics[width=1.0\linewidth]{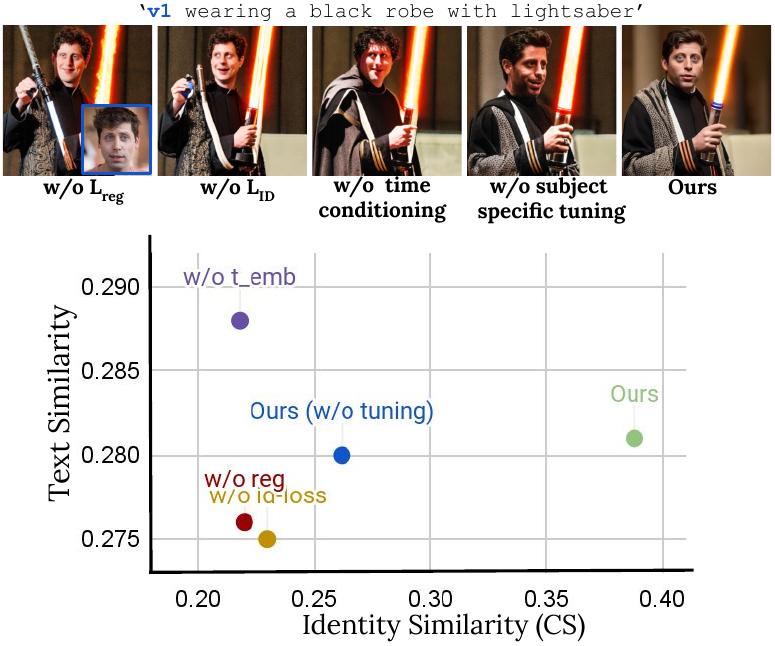}
    \vspace{-8mm}
    \caption{\textbf{Ablation study}} 
    \label{fig:quant-results-embed}
    \vspace{-8mm} 
\end{wrapfigure}  

\subsection{Ablations}
\label{subsec:ablations} 
We ablate over the design choices made in the proposed approach for personalization in Fig.~\ref{fig:quant-results-embed}. The Identity loss and regularization loss have a similar effect in pushing the token embeddings close to an embedding region for faces. Time-dependent token embedding is crucial to preserving subjects' identity as it provides a more expressive space to represent the face. Finally, subject-specific tuning with combined loss improves the Identity similarity as well as the Prompt similarity as the predicted token embeddings are pushed closer to the editable region with $\mathcal{L}_{reg}$ and $\mathcal{L}_{ID}$.

\section{Discussion}
\noindent \textbf{Conclusion.} We present a novel framework to condition T2I diffusion models on $\mathcal{W+}$ space of StyleGAN2 models for fine-grained attribute control. 
Specifically, we learn a latent mapper that projects the latent codes from $\mathcal{W+}$ to the input token embedding space of the T2I model learned with denoising, regularization, and identity preservation losses. This framework provides a natural way to embed a real face image by obtaining its latent code using the GAN encoders model. The embedded face can then be edited in two manners - coarse text-based editing and fine-grained attribute editing by latent manipulation in $\mathcal{W+}$.   

\noindent \textbf{Limitations.} 
The primary limitation is that the encoder-based inversion in $\mathcal{W+}$ is loose on some information hence we need to perform test time fine-tuning for a few iterations to recover identity similar to pivotal tuning. Additionally, the current approach utilizes multi-diffusion for composing multiple persons which requires multiple diffusion processes. Effectively composing more than two individuals with consistent identities proves challenging within the current method and is an interesting future direction to explore.


        

\clearpage

\noindent \textbf{Ethics statement.} We propose a face personalization method that might be useful for fake news generation as it enables the generation of persons in novel contexts. However, this issue is not limited to this work as it exists in several personalization approaches~\cite{celeb-basis,e4t} and generative models ~\cite{ldm}. Nonetheless, recent works~\cite{kumari2023ablating} have leveraged the power of the same personalization approaches for removing concepts or biases from T2I models.

\noindent \textbf{Acknowledgements.} 
We thank Aniket Dashpute, Ankit Dhiman and Abhijnya Bhat for reviewing the draft and providing helpful feedback. This work was partly supported by PMRF from Govt. of India (Rishubh Parihar) and Kotak IISc AI-ML Centre. 

%
%

\bibliographystyle{splncs04}
\bibliography{main}
\end{document}


\title{Supplementary Document - PreciseControl: Enhancing Text-To-Image Diffusion Models with Fine-Grained Attribute Control} 

\titlerunning{Abbreviated paper title}

\author{First Author\inst{1}\orcidlink{0000-1111-2222-3333} \and
Second Author\inst{2,3}\orcidlink{1111-2222-3333-4444} \and
Third Author\inst{3}\orcidlink{2222--3333-4444-5555}}

\authorrunning{F.~Author et al.}

\institute{Princeton University, Princeton NJ 08544, USA \and
Springer Heidelberg, Tiergartenstr.~17, 69121 Heidelberg, Germany
\email{lncs@springer.com}\\
\url{http://www.springer.com/gp/computer-science/lncs} \and
ABC Institute, Rupert-Karls-University Heidelberg, Heidelberg, Germany\\
\email{\{abc,lncs\}@uni-heidelberg.de}}

\maketitle 

\appendix

\section{Index}
In this supplementary document, we provide implementation details \textbf{(Sec.B)}. Next, we present ablation over different design choices \textbf{(Sec.C)} for mapper training, and subject-specific tuning \textbf{(Sec.D)}. We present additional comparison results for attribute editing \textbf{(Sec.E)} and face-restoration results \textbf{(Sec.F)}. We present editing results using InterfaceGAN edit directions \textbf{(Sec.G)}. Finally, we provide additional single-subject and multi-subject composition results \textbf{(Sec.H and I)} and comparison with other encoder-based inversion methods ~\textbf{(Sec.J)} followed by discussions on limitations \textbf{(Sec.K)} and prompts used \textbf{(Sec.L)}.  

\section{Implementations}
\subsection{Our Method}

\noindent \textbf{Latent Adapter} Latent adapter is a $4$-layer mapper MLP similar to StyleGAN mapper that maps the StyleGAN's $w$ latent code to clip token embedding space. To get more control over image generation, we generate different token embedding at each diffusion timestep by taking the timestep as input to the mapper. First, the $w$ latent is passed through a self-attention layer, and the output is flattened. Then, the timestep embedding is passed through a $2$-layer MLP after positional encoding, and the output is concatenated with the flattened feature from the self-attention layer, which is finally passed through a $4$-layer StyleGAN MLP model. During training, we kept $\lambda_{reg}=\num{1e-7}$ and $\lambda_{ID}=1.0$ in all the experiments as it achieves a good balance between identity preservation and text editability. We train the latent adaptor for $150K$ iterations with a batch size of $2$, which takes around $15-16$hrs on a single NVIDIA-$A6000$ GPU.

\noindent \textbf{Computation of ID loss.} We obtain a predicted clean image ($\hat{x_0}$) using $x_t$ at timestep $t$ using standard DDIM~\cite{ddim}. The loss is computed with the source image ($\mathcal{I}$) as $\mathcal{L}_{ID} = ||\mathcal{E}_{ID}(\mathcal{I}) - \mathcal{E}_{ID}(\hat{x_0})||^2$, where $\mathcal{E}_{ID}$ is the face recognition model~\cite{curr-face-rec}.

\noindent \textbf{LoRA Finetuning.} During LoRA fine-tuning, we train the latent mapper and LoRA weights for $50$ iterations with a learning rate of $\num{1e-3}$ for the LoRA weights and $\num{5e-6}$ for the latent mapper. Along with the standard diffusion loss, we add identity loss and regularisation loss for the loss calculation. The model used for identity loss is an Arcface ResNet $50$ variation~\cite{curr-face-rec}. 

\noindent \textbf{Edit Directions.} To have a unified framework to extract latent edit directions for all the attributes, we take a simple approach where we generate image pairs before and after editing using off-the-shelf editing methods and then aggregate the difference between the projected latent codes to obtain global edit direction. This enables us to obtain the edit direction of any new attribute by generating image pairs from a specialized model. 

\subsection{Baselines}
For a single-person generation, we compare our model with five baseline methods: Dreambooth~\cite{ruiz2023dreambooth}, Dreambooth with LoRA finetuning, Textual Inversion~\cite{textual-inversion}, Custom Diffusion~\cite{custom-diffusion} and CelebBasis~\cite{celeb-basis}. We use Stable Diffusion v2.1 for all methods. We use the official implementation and parameters of Celeb Basis. We use the diffusers' implementation for Custom Diffusion, Dreambooth, Dreambooth+LoRA, and Textual Inversion. These methods gave poor results with $5$ input images with a lower learning rate and higher training iterations to avoid overfitting. 
We train Dreambooth with a learning rate of 2e-6, batch size of $1$ for $800$ training steps with an instance prompt, "A photo of the face of a v1 person". For Dreambooth+LoRA, we use a learning rate of $\num{1e-4}$, batch size of $1$, for $1000$ training steps with the same instance prompt. Both dreambooth methods are trained with a train text encoder and prior preservation.
For Textual Inversion, we use an initializer token “face” with a batch size of $8$, and a learning rate of $\num{5e-3}$ for $5000$ steps.  
For custom diffusion, we train the model for $2000$ iterations with a batch size of $4$ and a learning rate of $5e-6$ and train all the parameters in the cross-attention layer as they have suggested. For multi-person cases, we train Custom Diffusion and Textual Inversion with $5$ images of each subject and a single image for each subject for ours and Celeb Basis. 

\section{Ablation over design choices} 
We ablate over different design choices of our method in Fig.~\ref{fig:supply-quant-ablation}. Regularization

\begin{wrapfigure}{r}{0.50\textwidth}
    \centering
    \includegraphics[width=0.95\linewidth]{figs/[sig2024]-supply-single-subject-ablate-quant.pdf}
    \vspace{-4mm}
    \caption{\textbf{Ablation study over different model components.}} 
    \label{fig:supply-quant-ablation}
    \vspace{-4mm} 
\end{wrapfigure}   

\noindent loss and Identity loss have a similar impact on personalization. Without these losses, the model performs poorly in both text similarity and identity similarity. Without time-step conditioning, the model has higher text editability as the predicted token embeddings are less expressive concerning subject identity and are close to the \textit{face} class token embeddings. However, it struggles with poor identity preservation. Having a subject-specific tuning improves the model's performance in both text editability and identity preservation.

\section{Ablation experiments for subject specific tuning.}
We experiment over the LoRA parameters for subject specific tuning in Fig.~\ref{fig:lora-training-iter-ablate} and Fig.~\ref{fig:lora-weight-ablate}.
In Fig.~\ref{fig:lora-training-iter-ablate} we change the number of iterations for LoRA training. Empirically, we found the model achieves good identity with $50$ training iterations. Increasing the training iterations does not improve the results. We ablate over the LoRA weight $\alpha$ used during inference in Fig.~\ref{fig:lora-weight-ablate}. Having a higher LoRA weight results in overfitting to the identity, and a lower weight results in inferior identity reconstruction. 

\begin{figure}[h]
    \centering
    \vspace{-5mm}
    \includegraphics[width=\linewidth]{figs/[sig2024]-supply_lora_iteration_abalation.pdf}
    \caption{\textbf{Ablate over training iterations for subject-specific tuning}. We found training for $50$ iterations is sufficient to achieve good identity reconstructions.}
    \label{fig:lora-training-iter-ablate}
    \vspace{-10mm}
\end{figure} 

\begin{figure}
    \centering
    \includegraphics[width=\linewidth]{figs/[sig2024]-supply_lora_weight_abalation.pdf}
    \caption{\textbf{Ablation over LoRA weight.} Having higher LoRA weight $\alpha$ leads to overfitting the identity of the subject and having poor text similarity. On the other hand with lower LoRA weights identity reconstruction is poor. We found the LoRA weight range between $0.2-0.4$ works best for most of the cases. Additionally, this provides a control knob to the tradeoff between identity preservation and prompt similarity.}
    \label{fig:lora-weight-ablate}
\end{figure}

\section{Comparison for editing with text-based editing method}
We compare our method for fine-grained attribute editing against text-based editing methods - Imagic~\cite{imagic}, and Instruct-pix2pix~\cite{brooks2023instructpix2pix}. Imagic is a single-image personalization method that fine-tunes the diffusion model to obtain inversion and then perform text-based edits.  Instruct-pix2pix is an image-to-image translation model trained on a large dataset for instruction-based editing. Additionally, we also compare with prompt-based editing on the learned subject by giving edit prompts such as \textit{`A v1 person smiling'}. The comparison results are present in Fig.~\ref{fig:supply-text-based-edit}. The proposed method achieves excellent editing results by changing only the subject's faces. Other approaches either change the background or pose of the subject during editing.  

\begin{figure}[h]
    \centering
    \includegraphics[width=0.90\linewidth]{figs/[sig2024]-supplimentary_full_body_edit_comparision.pdf} 
    \caption{\textbf{Attribute editing comparison.} We compare our proposed method with text-based editing methods Imagic and Instruct-pix2pix. We also compared with our personalization method but using text prompts for editing, e.g., \textit{`A v1 person smiling'}. Imagic can perform the edit but changes the clothes of the subject and background appearance. Instruct-pix2pix performs more disentangled edits but still changes the shirt of the subject in column $2$ and the background in column $4$. Prompt-based editing results in pose changes in the subject along with other changes. The proposed method with $\mathcal{W+}$ latent-based editing achieves the best performance and generates highly disentangled attribute edits, preserving the subject's appearance and background.}
    \label{fig:supply-text-based-edit}
\end{figure}

\clearpage

\section{Face restoration}

\begin{figure}[h]
    \centering
    \vspace{-10mm}
    \includegraphics[width=0.70\linewidth]{figs/[sig2024]-supply-face-restoration-outputs.pdf}
    \vspace{-8mm} 
    \caption{\textbf{Additional applications} - \textbf{(Top) Face restoration}: The disentangled $\mathcal{W+}$ latent space serves as an excellent face prior in StyleGANs. Leveraging this, we perform face restoration by projecting a corrupted face image into the $\mathcal{W+}$ latent space using the StyleGAN encoder to condition the T2I model. The resultant generated images look realistic and can be embedded and edited using T2I capabilities. Personalization of such corrupted images is challenging as the model could easily overfit the given single-face image.
    \textbf{(Bottom) Sketch-to-image}: We use psp~\cite{psp} encoder trained to map edge images to the $\mathcal{W+}$ latent space for sketch to image generation. The obtained $w$ latent code can then be used to condition the T2I model for real image generation.}
    \vspace{-30mm}
    \label{fig:face-restoration}
\end{figure}
\clearpage

\section{Results for InterfaceGAN edit direction}
To show the generalization of the edit directions from StyleGANs we use the attribute edit directions obtained from InterfaceGAN to perform editing in Fig.~\ref{fig:interface-edits}. The proposed method can generate disentangled attribute edits suggesting the editing direction obtained by off-the-shelf StyleGAN editing method could be used for editing in T2I models. 

\begin{figure}
    \centering
    \includegraphics[width=\linewidth]{figs/supp_other_edit_directions.pdf}
    \caption{\textbf{Editing with InterfaceGAN directions.} We perform attribute editing by using the edit directions obtained from InterfaceGAN~\cite{interface}. The edit directions generalize well in T2I models, indicating easy integration of any off-the-shelf StyleGAN editing method in our framework.}
    \label{fig:interface-edits}
\end{figure} 

\clearpage

\section{Comparison with multi-subject personalization}
We compare our method for the task of multi-person composition with Custom Diffusion~\cite{custom-diffusion} and FastComposer~\cite{xiao2023fastcomposer} in Fig.~\ref{fig:supply-multi-person-compare}. FastComposer performs extensive training on the T2I model on images with multiple people and the corresponding text prompts. This allows them to compose multiple subjects during inference. Both ours and Custom Diffusion perform subject-specific fine-tuning to embed a given subject. 

\begin{figure}
    \centering
    \vspace{-7mm}
    \includegraphics[width=0.9\textwidth]{figs/[sig2024]-supply-multi-person-compare.pdf}
    \caption{\textbf{Comparison for multi-person personalization.} We compare two state-of-the-art personalization methods for multiple concepts FastComposer~\cite{xiao2023fastcomposer} and Custom Diffusion~\cite{custom-diffusion} FastComposer has an extensive pretraining stage where the T2I models are fine-tuned with images of multiple subjects and text. Custom Diffusion is able to generate two persons following the text prompt; however, it suffers from \textit{attribute mixing} where the same subject is generated twice or facial features are mixed. FastComposer can better preserve the identity but the composition always has two closeup faces in the scene and does not follow the layout given by the text prompt. Our proposed method attains excellent identity preservation and follows the text prompt during generation.} 
    \vspace{-10mm}
    \label{fig:supply-multi-person-compare}
\end{figure}

\clearpage

\section{Comparison results for single subject personalizaton}
We present additional comparison results against the same set of personalization methods Dreambooth, Dreambooth+LoRA Textual Inversion, Custom Diffusion, Celeb Basis in Fig.~\ref{fig:supply-single-person-compare}. We can observe our method is able to generate highly realistic face images that follow the text prompts. Additionally, our method achieves a good balance between Identity similarity and Text Editability. 

\begin{figure}[h]
    \centering
    \vspace{-5mm}
    \includegraphics[width=0.8\textwidth]{figs/[sig2024]-supply-single-person-compare.pdf}
    \vspace{-2mm}
    \caption{\textbf{Single person personalization.} We compare against state-of-the-art personalization methods. Custom Diffusion achieves excellent identity preservation by generating closeup faces but has lower prompt similarity (row 5 \& 6). Dreambooth~\cite{ruiz2023dreambooth}, Dreambooth+LoRA, and Textual Inversion~\cite{textual-inversion} improve text editability but fail to reconstruct the identity. Celeb-Basis~\cite{celeb-basis}, a face-specific personalization approach, has better identity preservation and text alignment, but the faces have an unnatural cartoonish look. Our method achieves a good tradeoff and generates highly realistic face images.}
    \vspace{-6mm}
    \label{fig:supply-single-person-compare}
\end{figure} 

\clearpage

\clearpage

\clearpage

\section{Additional baseline comparisons}
We compare our proposed method with recent encoder-based personalization methods in Fig.~\ref{fig:encoder-qual-comparision} and Fig.~\ref{fig:encoder-compare}. To demonstrate the robustness of our method across datasets, we curated two new datasets that are potentially unseen by the diffusion models during training for fairness.

\begin{enumerate}
    \item \textbf{Synthetic-faces [SF]} - We create a dataset of $30$ individuals generated from pretrained StyleGAN2 on FFHQ~\cite{sg} dataset. 
    \item \textbf{Real-faces [RF]} - We extract frames from RAVDESS~\cite{ravdess} video emotion recognition dataset. This dataset contains videos of $24$ individuals acting different expressions captured in a lab environment. We extract one frame per individual from these videos to create a dataset of real faces. We don't use popular datasets scraped from the web because they can potentially be in the training set of StableDiffusion~\cite{ldm}. 
\end{enumerate}

\begin{figure}
    \centering
    \includegraphics[width=0.8\linewidth]{figs/[sig2024]-supply-sd15-qual-comparision.pdf}
    \caption{Qualitative comparison of Encoder-based models with ours, and for a fair comparison, we have downgraded our model and trained it on SDv1.5. But our original method is in SDv2.1, and it has significantly better results Fig.\textcolor{red}{7},\textcolor{red}{8},\textcolor{red}{13},\textcolor{red}{14}.}
    \label{fig:encoder-qual-comparision}
\end{figure}

\noindent We compare with state-of-the-art encoder based methods - \textbf{IP-Adapter}~\cite{ipadaptor}, \textbf{E4T}~\cite{e4t} and $\mathcal{W}+$\textbf{adaptor}~\cite{sg2diff} on the above two datasets of unseen individuals (SF \& RF). The open-source code for these baselines is on the previous version of Stable Diffusion (SD); we implement our method on SDv1.5 for a fair comparison. Our method outperforms encoder models in Prompt-similarity (PS) and ID preservation (CS). Celeb-basis and $\mathcal{W}$+adaptor achieve high prompt alignment; however, they suffer in preserving subjects' identity. On the contrary, E4T, IP-Adapter, and our approach have improved identity preservation. Our proposed method achieves the best tradeoff in ID preservation and prompt similarity. Note, all the encoder methods except $\mathcal{W}$+adapter (to which we compare next) \textit{cannot perform fine-grained attribute editing} and rely only on coarse text-based editing. Our method enables smooth control over attribute edits along with exceptional identity preservation during personalization while preserving text-based editing.

\begin{figure}
    \centering
    \includegraphics[width=0.8\linewidth]{figs/supply_encoder_compare.pdf}
    \caption{Comparison with state-of-the-art encoder-based personalization methods on real and synthetic face datasets.} 
    \label{fig:encoder-compare}
\end{figure}

\begin{figure}
    \centering
    \includegraphics[width=0.8\linewidth]{figs/[sig2024]-supply-wadapter-img_gen-comp.pdf}
    \caption{Comparison with $\mathcal{W}$+adapter for different text prompts and identities. For each identity, the first row is $\mathcal{W}$+-adapter's, and the second row is our method result. $\mathcal{W}$-adapter is not able to reconstruct the subject with accurate identity, whereas out method generates realistic images with superior identity preservation.} 
    \label{fig:wadapter_generation_comp}
\end{figure}

\begin{figure}
    \centering
    \includegraphics[width=0.8\linewidth]{figs/[sig2024]-supply-wadapter-edit-comp.pdf}
    \caption{Fine-grained attribute editing comparison with $\mathcal{W}$+adapter. For each identity, the first row is $\mathcal{W}$+adapter, and the second row is our method result. 
    Edits generated by $\mathcal{W}$+adapter entangle multiple face attributes. In most results, the background is significantly changed with edits (example 2 \& 4). Our proposed method generates highly disentangled edits while preserving backgrounds. Here, we can observe entanglement and background change since there are no constraints during the image generation process to fix the image layout.} 
    \label{fig:wadapter_edit_comp}
\end{figure}

\vspace{-2mm}
\subsection{Comparison with $\mathcal{W}$+adaptor.}
A concurrent work $\mathcal{W}$+adaptor~\cite{sg2diff} aimed to condition the text-to-image models with $\mathcal{W+}$ latent space of StyleGAN2 for controlling the generation process. We want to highlight the key differences between $\mathcal{W}$+adaptor along with results in Fig.~\ref{fig:wadapter_generation_comp},~\ref{fig:wadapter_edit_comp}.
Firstly, $\mathcal{W}$+adaptor trained a mapping network to map the $w$ latent code to a set of latent features $f_z$. The latent features $f_z$ are then injected along with the text embeddings using a residual cross-attention to embed a face. Our method takes a more effective approach of inverting a face directly in the token embedding space while being close to the \textit{`person'} concept in the embedding space via regularization. This ensures good fidelity inversion and consistency in subject identity during generation across different prompts, whereas $\mathcal{W+}$adapter generates inconsistent faces with different prompts (Fig.~\ref{fig:wadapter_generation_comp}). In the second and third examples, we can clearly see that identity is changing across different prompts. Further, StyleGAN2 encoder-based inversion~\cite{e4e} in $\mathcal{W}+$ inherently results in identity loss, which is also observed in $\mathcal{W}$+adaptor results (Fig.~\ref{fig:wadapter_generation_comp}). Specifically, for Einstein, the identity is distorted during inversion. Our proposed subject-specific tuning (of only $50$ iterations) overcomes this issue and recovers the identity while preserving rich latent properties from $\mathcal{W+}$ space. Additionally, we compare latent-based attribute editing results with $\mathcal{W}+$adapter in Fig.~\ref{fig:wadapter_edit_comp}. During editing, we observe a high entanglement with the background and other facial attributes. Specifically, in the second example, we observe gender changes along with the beard edit; similarly, for the third example, we can see that pose and face size change along with the race attribute. Finally, $\mathcal{W+}$-adaptor fails to generate compositions of more than one subject. In contrast, our method supports the generation of multiple objects and allows for selective latent-based editing of individual subjects (Fig.\textcolor{red}{1},\textcolor{red}{6}).

\clearpage

\begin{wrapfigure}{r}{0.50\textwidth}
    \centering
    \vspace{-8mm}
    \includegraphics[width=0.95\linewidth]{figs/supply-limitations-threeperson.pdf}
    \vspace{-2mm}
    \caption{\textbf{Failure case}} 
    \label{fig:three-person-failure}
    \vspace{-10mm} 
\end{wrapfigure}   

\section{Limitations.}
Though our method performs well for personalization, it has the following limitations: 

\begin{itemize} 
    \item We use multiple diffusion processes for generating compositions. Although one can parallelize these processes, it increases inference time. 
    \item Our method fails to accurately generate the composition of three or more subjects, as shown in Fig.~\ref{fig:three-person-failure}, which is an interesting future research direction.
\end{itemize} 

\vspace{-6mm}
\begin{itemize}
    \item Existing limitations of StyleGAN2 (entanglement at higher edit strengths \& sequential-editing) editing also leaks in our editing framework.
    \item As we use delayed conditioning for better text editability, the initial seed determines image layout, which can cause some bad layouts for some seeds; this is more relevant in multiple-person scenes, where T2I has trouble generating multiple persons.  
\end{itemize}

\section{Text prompts for generation} 

\subsection{Single person prompts}
\begin{itemize}
    \item Manga drawing of sks person
    \item Ukiyo-e painting of sks person
    \item Cubism painting of sks person
    \item Banksy art of sks person
    \item Cave mural depicting of sks person
    \item sand sculpture of sks person
    \item Colorful graffiti of a sks person
    \item Watercolor painting of sks person
    \item Pointillism painting of sks person
    \item sks person stained window glass
    \item sks person latte art
    \item sks person as a greek sculpture
    \item sks person as a knight in plate armour
    \item sks person wears a scifi suit in space
    \item sks person in a chinese vegetable market
    \item sks person in a superhero costume
    \item sks person on the beach
    \item sks person is playing the guitar
    \item sks person eats bread in front of the Eiffel Tower
    \item sks person wearing yellow jacket, and driving a motorbike
    \item sks person wearing a santa hat in a snowy forest
    \item sks person wearing a black robe with a light saber in hand
    \item sks person as a firefighter in front of a burning building
    \item sks person camping in the woods in front of a campfire
    \item sks person taking a selfie in front of grand canyon
\end{itemize}

\subsection{Multi person prompts} 

\begin{itemize}
    \item A painting of sks person and ks person in van gogh style
    \item A sks person and ks person taking photo after a tennis game
    \item A sks person and ks person play the piano together
    \item sks person and ks person on the beach
    \item sks person and ks person sitting on a park bench
    \item sks person and ks person eating a gaint pizza in a restaurant
    \item A sks person and ks person having dinner together
    \item A sks person and ks person playing chess, closeup
    \item A sks person and ks person wearing a Christmas hat in a snowy forest
    \item A sks person and ks person sitting on sofa
    \item A sks person and ks person baking a cake together
    \item sks person and ks person discussing a math problem intensly
    \item sks person and ks person fighting in a ring
    \item sks person and ks person singing karaoke together in a bar, colorful lights
    \item sks person and ks person in a colorful bar
    \item sks person and ks person accepting an award on stage
\end{itemize}

\clearpage

%
%
\bibliographystyle{splncs04}
\bibliography{main}